\documentclass[journal]{IEEEtran}
\usepackage{multirow} 
\usepackage{booktabs}
\usepackage{makecell}
\usepackage{cite}
\usepackage{amsmath} 
\usepackage[ruled]{algorithm2e} 
\usepackage{array}
\usepackage{graphicx}
\usepackage{float}
\usepackage{amsthm,amsmath,amssymb}
\usepackage[table,xcdraw]{xcolor}
\graphicspath{{./figures/}}
\usepackage{url}  

\ifCLASSOPTIONcompsoc
  \usepackage[caption=false,font=normalsize,labelfont=sf,textfont=sf]{subfig}
\else
  \usepackage[caption=false,font=footnotesize]{subfig}
\fi

\begin{document}
\title{A Label-Free High-Precision Residual Moveout Picking Method for Travel Time Tomography based on Deep Learning}
\author{Hongtao~Wang, Jiandong~Liang, Lei~Wang, Shuaizhe~Liang, Jinping~Zhu, Chunxia~Zhang, Jiangshe~Zhang.
\thanks{Corresponding authors: Chunxia~Zhang, Jiangshe~Zhang}
\thanks{H.T. Wang, C.X. Zhang, J.S. Zhang are with the School of Mathematics and Statistics, Xi'an Jiaotong University, Xi'an, Shaanxi, 710049, P.R.China.}
\thanks{J.D. Liang, L. Wang, S.Z. Liang, J.P. Zhu are with China National Petroleum Corp Bureau of Geophysical Prospecting Inc, Geophysical Technology Research Center Zhuozhou, Hebei, 072750, P.R.China.}
\thanks{The research is supported by the National Natural Science Foundation of China under grant 12371512; and in part by the Key Acquisition, Processing and Interpretation Techniques for Seismic Data Processing of China National Petroleum Corporation (CNPC) under Grant 01-03-2023.}}

\markboth{Journal of \LaTeX\ Class Files,~Vol.~xx, No.~x, Dec~2024}%
{Wang \MakeLowercase{\textit{et al.}}: A DL-based High-Precision Residual Moveout Picking Method}

\maketitle

\begin{abstract}
Residual moveout (RMO) provides critical information for travel time tomography.
The current industry-standard method for fitting RMO involves scanning high-order polynomial equations. However, this analytical approach does not accurately capture local saltation, leading to low iteration efficiency in tomographic inversion.
Supervised learning-based image segmentation methods for picking can effectively capture local variations; however, they encounter challenges such as a scarcity of reliable training samples and the high complexity of post-processing.
To address these issues, this study proposes a deep learning-based cascade picking method. It distinguishes accurate and robust RMOs using a segmentation network and a post-processing technique based on trend regression. Additionally, a data synthesis method is introduced, enabling the segmentation network to be trained on synthetic datasets for effective picking in field data. Furthermore, a set of metrics is proposed to quantify the quality of automatically picked RMOs.
Experimental results based on both model and real data demonstrate that, compared to semblance-based methods, our approach achieves greater picking density and accuracy.
\end{abstract}
\begin{IEEEkeywords}
Residual moveout picking, Depth domain velocity modeling, Deep learning.
\end{IEEEkeywords}

\IEEEpeerreviewmaketitle

\section{Introduction}

\IEEEPARstart{T}{omographic} inversion is extensively utilized in geophysics, significantly advancing resource exploration \cite{brzostowski19923}. It also plays a crucial role in disaster control\cite{dou2012rockburst}, environmental protection\cite{Colin2026groundwater}, and engineering safety\cite{wang2017mechanism}. Residual moveout (RMO) provides essential information for travel time tomography, allowing for the updating of the initial velocity model \cite{woodward1998automated}. Accurate RMO picking is crucial for tomographic velocity model building. Specifically, RMO establishes a flatness criterion for events within a common image gather (CIG) of both the offset domain or angle domain \cite{adler2008nonlinear}. Any deviation of the focused reflection or diffraction signals in a CIG from a flat configuration serves as a quantitative indicator of inaccuracies in the velocity model. Consequently, the RMO can be utilized as input for the travel time tomography inversion, facilitating the refinement of the migration velocity model. However, extracting RMO (equivalent to residual curvatures) from CIG is huge for manual effort. The RMO picking had to be performed automatically.

The residual curvatures in GIGs usually present parabola or higher-order polynomial curve. Consequently, conventional RMO picking methods are based on coherence and semblance scans \cite{AlYahy1989semblance} method, where the velocity parameters are optimized using scanning and then the updated velocity model applies normal moveout (NMO) correction to flatten the CIGs. However, the scanning-based methods need perform several iterations, which requires intensive labor and high computational cost \cite{Audebert1997scan,Woodward1998scan}. Aiming to pick RMO for a three-dimensional survey, Adler and Brandwood \cite{Adler1999} first implemented a sparse semblance scan and then conducted a locally scaled regression to ensure a robust and smooth picking. In case of far-offset CIGs, the residual curvature becomes complex, a higher degree of freedom is required to approximate it, e.g., a higher-order polynomial equation or a scan method with more optimal parameters. Compared with conventional one-parameter scan method, Siliqi et al. \cite{siliqi2003high} performed a velocity-anellipticity scan to approximate non-hyperbolic moveout in terms of large offsets and steep dips. Subsequently, aiming to resolve the anisotropy case, an uncorrelated scanning with multi-parameter are proposed \cite{siliqi2007high}. Moreover, a semi-automatic pickup based on the estimation of fourth-order cumulants was developed to guarantee the accuracy for the late tomographic iterations\cite{zhang2014automatic}. With the increase in the order of the equation, the computational cost rises significantly. To improve the optimization efficiency of high-order scan methods, Xu and Zhang \cite{xu2024high} proposed a high-order RMO estimation method that incorporates global optimization within local time windows. 
However, there is a trade-off between the degree of freedom in the descriptive equation and the computational cost associated with the parametric approach. Nonparametric methods can eliminate the constraints of assumptions and effectively characterize local anomalous changes \cite{Liu2010rtm,fruehn2014resolving}.
Unfortunately, while nonparametric methods can effectively describe locally rapid variations, they are not robust in low signal-to-noise ratio CIGs. Furthermore, nonparametric methods estimate each RMO locally, which also incur significant computational costs. 



With the advancement of machine learning (ML) and the increasing demand for higher accuracy in RMO picking, various ML-based methods for RMO picking have been developed.
Since it is difficult for semblance-based methods to control the quality of RMO pickings \cite{Harris1999uncertainty}, Zhou and Brown \cite{Zhou2020mlpicking} utilized a random forest classifier and a neural network classifier to identify whether the semblance-based pick is a good picking or not. However, this correction post-processing only achieves removing the bad picking results, but not providing accurate pickings. 
Compared with ML-based post-processing method, Bazargani et al. \cite{Bazargani2022segmentation} proposed a novel ML-based multi-stage RMO picking framework. Specifically, a semantic segmentation network first infers a binary segmentation image based on CIG, where the pixels closed to 1 in the image indicate the residual curvatures. Then, a hierarchical clustering method is conducted to split each curvatures from the segmentation map. Finally, similar with Adler and Brandwood \cite{Adler1999}, a robust fitting algorithm is implemented to estimate smooth residual curvatures. Semantic segmentation is not only used in RMO picking, but also in similar tasks such as horizon detection \cite{Tschannen2020horizon}, fault detection \cite{Xiong2018fault} and first break picking \cite{Wang2024FB}. Although segmentation-based methods can capture complex curvatures, additional post-processing is essential after segmentation. Furthermore, the number of curvatures required during post-processing necessitates extra human intervention.
Additionally, all of above ML-based methods are supervised-learning methods, which depend on adequate training datasets with manual annotation to support the training processing of model weights. To tackle this issue, Wu et al. \cite{Wu2019fault} proposed a novel generation method of synthetic fault datasets and predicted the field data successfully based on the model trained on synthetic training datasets. As best as we know, there is no synthetic data set generation method in the task of RMO picking, and accurate manual annotation of RMO picking task is extremely difficult, so there is no open source data set for scholars to study. 
Aiming to fundamentally solve the problem of labeling data set dependence, reinforcement learning (RL)-based approach avoids the labeling-based training process. Wu et al. \cite{Wu2023RL} applied Markov decision process (MDP) to pickup the residual curvature point-wise based on a assigned start point, where a reward function with multiple seismic attributes (i.e., waveform and instantaneous phase) is defined to guide the picking strategy between the adjacent traces. Similarly, RL-based tracking methods can be transferred to first break task, which define a different reward function to fit the new scene \cite{Ma2018RL,Ma2019RL}. Although MDP-based RMO picking methods avoid the requirement of the labeled datasets, choice of start points and the robustness in case of multiple noise still obstruct the accuracy of the field picking.



In this article, we propose a novel cascade framework to pick robust residual curvatures in CIGs, where a semantic segmentation first recognizes residual curvatures based on the images of CIGs, and each curvature is then extracted through a post-processing step that is primarily guided by slope field constraints. Unlike the multi-stage picking framework \cite{Bazargani2022segmentation}, our approach addresses three key challenges:
\begin{itemize}
\item We propose a novel method for generating synthetic CIGs with residual curvatures, addressing the shortage of annotated datasets.
\item The proposed post-processing method does not require manually assigned curvature numbers and maintains a consistent trend in the picked curvatures.
\item The quantitative index for RMO picking is inadequate. Our work defines two metrics to evaluate RMO picking based on manual picking methods.
\end{itemize}

The structure of this article is summarized as follows: Section II details our proposed cascade picking framework. Subsequently, Sections III and IV introduce the generation method for synthetic CIGs and the new metrics for RMO picking, respectively. Section V evaluates our method using both model and field datasets. Furthermore, Section VI analyzes the effectiveness of key components and new metrics, as well as the sensitivity of inference hyperparameters. Finally, Section VII concludes the paper.

\section{A Cascade Picking Framework for RMO Picking}

\subsection{Overview of Cascade Method}
\begin{figure*}[!ht]
    \centering
    \includegraphics[width=5in]{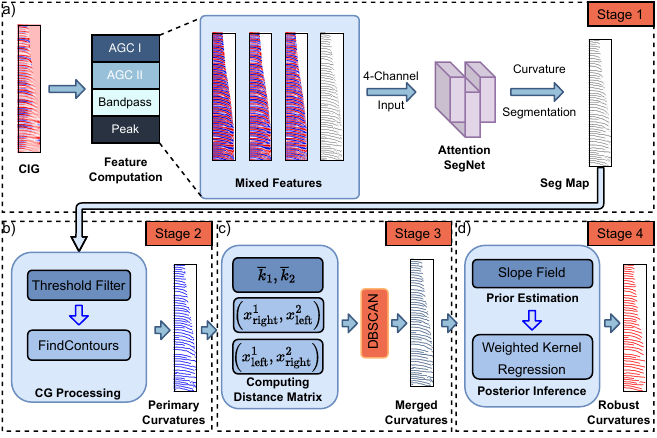}
    \caption{The four stages of our proposed cascade RMO picking framework.}
    \label{fig: mainflow}
\end{figure*}

The proposed cascade RMO picking framework implements the technique combination of the semantic segmentation and the slope-field constraint post-process. Generally, our proposed method includes four stages, and each stage is implemented in sequence. Fig. \ref{fig: mainflow} displays the flow chart of our proposed cascade method. Concretely, Stage 1 extracts mixed features from a CIG and then inputs attention segmentation network to infer a curvature segmentation map. Subsequently, Stage 2-4 achieves a post-process dominated by slope field constraints to draw out each curvature from the segmentation map. Initially, Stage 2 implements a computer graphic processing to find the contours of each curvatures. Then, Stage 3 concatenates the similar curvatures using a clustering method. Finally, Stage 4 performs a Bayesian inference to estimate the robust RMO picking. 

\subsection{Multi-Scale Attention Segmentation Network}
\subsubsection{Multi-Scale Feature Extraction}
The curvature feature in the raw CIG is not obvious, and it is necessary to enhance the texture information using signal processing methods. 
Since the amplitude scale of different regions in a CIG are various, we first adapt trace-wise automatic gained control (AGC) to normalize the local amplitude distribution:
\begin{equation}\label{eq: AGC-E}
E[t] = \frac{1}{h} \sum_{i=t-h+1}^{t} (d^{\text{time}}[i])^2,
\end{equation}
\begin{equation}\label{eq: AGC-NG}
d_{\text{AGC}}[t] = \frac{d^{\text{time}}[t]}{\sqrt{E(t)} + \epsilon},
\end{equation}
where $d_{\text{AGC}}[t]$ represents the AGC value of the time window with the length $h$ from $(t-h+1)$th - $t$th time samples, and $\epsilon$ is a positive value extremely close to 0. The AGC feature map concatenates all AGC vectors computed by each trace denoted as matrix $\mathbf{g}^{\text{AGC}}$. Moreover, the optimal hyper-parameter (window length $h$) of layers of different depth are usually distinct in a CIG, because of various dominant frequencies. Therefore, we select two window sizes $h_1$ and $h_2$ to compute two maps $\mathbf{g}^{\text{AGC}_1}$ and $\mathbf{g}^{\text{AGC}_2}$, respectively.

CIG also contains both low-frequency and high-frequency noise. Consequently, we implement a 1-dimensional (1-D) band-pass (BP) filtering based on 1-D fast fourier transformation (FFT) to eliminate the drastic frequencies. Under the discrete form, we define a allowed frequency range $[f_{\text{min}}, f_{\text{max}}]$, and mute the partial frequency out of the range in the frequency domain. Subsequently, the inverse FFT is preformed to cover the time-domain signal. The filtered time-domain signal vectors are concatenated to a feature map, denoted as matrix $\mathbf{g}^{\text{BP}}$. 
The residual curvatures in a CIG always locate or are close to the local peaks. Thus, we mask all local peaks in a map with the same shape as the CIG and denote the map as matrix $\mathbf{g}^{\text{Peak}}$. Finally, the four feature maps can be concatenated, forming a 4-channel tensor $\mathbf{g}^{\text{mix}}$ as the input of the segmentation network. 

\begin{equation}\label{eq: mix_feature}
\mathbf{g}^{\text{mix}} \triangleq [\mathbf{g}^{\text{AGC}_1}, \mathbf{g}^{\text{AGC}_2}, \mathbf{g}^{\text{BP}}, \mathbf{g}^{\text{Peak}}]
\end{equation}

\subsubsection{Information-Fusion Segmentation Network}
\begin{figure}[!ht]
    \centering
    \includegraphics[width=3in]{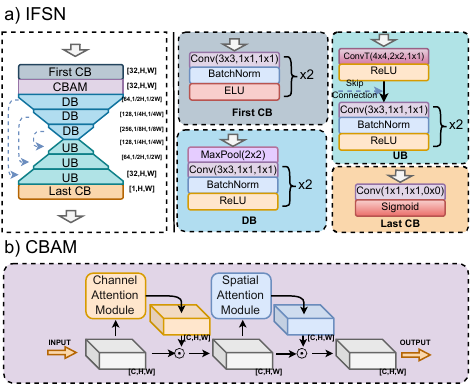}
    \caption{Flowchart of IFSN (a) and Details of CBAM (b).}
    \label{fig: IFSN}
\end{figure}

Abundant feature maps can provide sufficient information for segmentation network. At the same time, the error regions also can inject the noise to impede inferring curvatures of network. Consequently, aiming to use the multi-information features reasonably, we adapt an attention-based segmentation network. CBAM (Convolutional Block Attention Module) \cite{woo2018cbam} is outstanding technique for multi-information fusion, where both the channel attention and the spatial attention are preformed. As shown in Fig. \ref{fig: IFSN}(b), the shapes of the input and output of CBAM are coincident, making it easier to embed the attention module in the segmentation network. Concretely, the computation processing can be defined by:
\begin{equation}\label{eq: CBAM}
  \begin{array}{c}
    \mathbf{g}' = F_c(\mathbf{g})\odot \mathbf{g}, \\
    \mathbf{g}'' = F_s(\mathbf{g}')\odot \mathbf{g}',
  \end{array}
\end{equation}
where $F_c(\cdot)$ and $F_s(\cdot)$ are the functions of the channel attention module and the spatial attention module, and $\odot$ denotes element-wise (Hadamard) product.

In our work, we incorporate a CBAM into the classic semantic segmentation network, UNet \cite{ronneberger2015u}, to better handle ambiguous inputs. 
Proposed new network is named information-fusion segmentation network (IFSN), illustrated by Fig. \ref{fig: IFSN}(a). Concretely, the input of IFSN is a multi-information feature tensor with the shape of $4\times H\times W$, and the output of IFSN is the segmentation map of the residual curvatures with the shape of $H\times W$. The UNet used in this study includes three down-sampling blocks (DBs), three up-sampling blocks (UBs), and three special convolutional blocks (CBs). Specifically, CBAM is integrated at the end of the first CB to effectively fuse the information, as the CBAM includes a down-sampling operation for the channels. However, the number of input channels is insufficient for down-sampling. The spatial attention module and the channel attention module depicted in Fig. \ref{fig: IFSN}(b) adhere to the configuration outlined in the paper \cite{woo2018cbam}. Particularly, the spatial kernel size is set to 7, and the reduction factor is set to 32. Additionally, the output shapes of each layer are annotated on the right side of the IFSN in Fig. \ref{fig: IFSN}(a), which indicate the kernel numbers of each convolution layer.
Generally, The final output is a segmentation map with the same shape as the input CIG:
\begin{equation}\label{eq: net_output}
  \mathbf{g}^{\text{seg}} = F_{\text{IFSN}}(\mathbf{g}^{\text{mix}}),
\end{equation}
where $F_{\text{IFSN}}(\cdot)$ represents the function of IFSN.

\subsection{Trend-based Post-processing}
There is a gap between the segmentation map and the RMO. Consequently, we propose a novel post-processing method to extract curvatures from the segmentation using three stages. Fig. \ref{fig: mainflow} illustrates how these three stages achieve the splitting of preliminary curvatures, the merging of curvatures, and the estimation of robust curvatures, respectively. The following three subsections will detail each stage.

\subsubsection{Computer Graphics-based Curvature Split Method}
The pixels with high values in the segmentation map $\mathbf{g}^{\text{seg}}$ imply the semantic information of curvature. Therefore, the step processing in our post-processing is removing the noise pixels with the values lower than a threshold $T_{\text{seg}}$ and performing binaryzation:
\begin{equation}\label{eq: mute_seg}
{\mathbf{g}'}^{\text{seg}}[i, j] = \left\{
\begin{array}{ll}
  0 & \text{if}\ {\mathbf{g}}^{\text{seg}}[i, j]< T_{\text{seg}}, \\
  1 & \text{else},
\end{array}
\right.
\end{equation}
where ${\mathbf{g}}^{\text{seg}}[i, j]$ indicates the predict value of $i$th row, $j$th column pixel of the segmentation map ${\mathbf{g}}^{\text{seg}}$.

To distinguish each curvature in ${\mathbf{g}'}^{\text{seg}}$, we adapt a computer graphics-based (CG) method. Specifically, a counter-search algorithm\cite{suzuki1985topological} is implemented to retrieve outlines from a binary segmentation map ${\mathbf{g}'}^{\text{seg}}$. The output of the counter-search algorithm are a few discrete point sets of counters $\{\mathcal{C}_k\}_{k=1}^{K_1}$, where $K_1$ is the number of the found counters.
For each contour point set $\mathcal{C}_k$, we construct a polygon with each point in the set as a vertex and generate a mask $\mathbf{M}_k$ of the same size as ${\mathbf{g}'}^{\text{seg}}$, where the pixels inside the polygon are 1 and the contour and outside are 0. 

Finally, the preliminary curvatures can be computed by performing Hadamard product respectively: 
\begin{equation}\label{eq: Hadamard}
\mathbf{g}^c_k \triangleq \mathbf{g}^{\text{seg}} \odot \mathbf{M}_k, k=1,...,K_1,
\end{equation}
where $\mathbf{g}^c_k$ represents the mask map of $k$th curvature. Subsequently, the pixels with the maximum in each rows of $\mathbf{g}^c_k$ are regarded as the preliminary RMO picking:
\begin{equation}\label{eq: pre_RMO}
\mathbf{L}^1_k \triangleq \{\mathbf{x}_j\}_{j=1}^{n^1_k},
\end{equation}
where $\mathbf{x}_j$ is the $j$th point in the curvature $\mathbf{L}^1_k$, and $n^1_k$ is the point number of the $k$th preliminary RMO picking. Particularly, since the values of a few columns in $\mathbf{g}^c_k$ are all zero, these columns can not be considered in Eq. \ref{eq: pre_RMO}.

\subsubsection{New Distance Definition-based Clustering Method}
Stage 2 splits each curvature initially, but some curvatures should be merged, which belong to the same curvature realistically. Consequently, Stage 3 perform a new clustering method to merge these curvatures without superfluous manual control. We first define a novel distance between two curvatures, taking into account the mean slope as well as the left and right endpoints of each curvature. Concretely, the mixed distance equation $d^{\text{mix}}(\cdot, \cdot)$ are defined by:
\begin{equation}\label{eq: dist_mix}
d^{\text{mix}}(\mathbf{L}_1, \mathbf{L}_2) = \alpha * |s_1 - s_2| + (1-\alpha) * d^*(\mathbf{L}_1, \mathbf{L}_2),
\end{equation}
where $s_1$ and $s_2$ are the mean slopes of the curvatures $\mathbf{L}_1$ and $\mathbf{L}_2$, respectively, and $\alpha$ is a weight parameter to balance two parts. $d^*(\cdot, \cdot)$ computes a minimum distance between the endpoints of two curvatures:
\begin{equation}\label{eq: dist_new}
d^*(\mathbf{L}_1, \mathbf{L}_2) = \min\{\|(\mathbf{x}_1^{\text{right}}-\mathbf{x}_2^{\text{left}})\cdot \mathbf{w}\|_2, \|(\mathbf{x}_1^{\text{left}}-\mathbf{x}_2^{\text{right}})\cdot \mathbf{w}\|_2\},
\end{equation}
where $\mathbf{x}_1^{\text{left}}$ and $\mathbf{x}_1^{\text{right}}$ represent the left and right endpoints of $\mathbf{L}_1$, and the same is defined for $\mathbf{L}_2$. Additionally, $\mathbf{w}$ is a two-dimensional anisotropic weight vector used to assign weights to the offset and depth dimensions. In this work, we increase the influence of the offset dimension by setting $\mathbf{w} = [1, 1.5]^T$.
Particularly, the mean slope $s_k$ for curvature $L_k$ are averaged by the local slopes $s^{\text{local}}$, which is estimated window-by-window using:
\begin{equation}\label{eq: linear_reg}
s^{\text{local}} = \frac{n \sum_{i=1}^{n} o_i d_i - \sum_{i=1}^{n} o_i \sum_{i=1}^{n} d_i}{{n} \sum_{i=1}^{n} {o_i}^2 - (\sum_{i=1}^{n} o_i)^2},
\end{equation}
where $\{(o_i, d_i)\}_i$ is a point set of a local window, which is cropped from a curvature. 

For $K$ curvatures, a distance matrix $D$ with the shape of $K \times K$ are calculated using Eq. \ref{eq: dist_mix}. Aiming to merge a few curvatures based on $D$, we adapt a density-based clustering method, named DBSCAN (Density-Based Spatial Clustering of Applications with Noise) \cite{ester1996density}. There are two hyper-parameters in DBSCAN. One is the maximum distance between two samples for one to be considered, denoted as $d^{\text{eps}}$. The other is the number of samples in a neighborhood for a point $n^{\text{min}}$. 
It is found experimentally that the optimal $d^{\text{eps}}$ is related to the minimum interval of curvatures assumed in the field data, which is set to 4 in this work, while $n^{\text{min}}$ is set to 1, since some curvatures do not need to be merged. After the clustering process, we denote the merged curvatures as $\{\mathbf{L}^2_{k}\}_{k=1}^{K_2}$.

\subsubsection{Robust Bayesian Regression Method}
Although Stage 2 and 3 have split relatively unbroken curvatures, there are local high-frequency twists, which affect the tomography inversion adversely. Consequently, we propose a robust Bayesian regression method, aiming to refine the local unreasonable trends of $\{\mathbf{L}^2_{k}\}_{k=1}^{K_2}$ based on the global slope field. The proposed regression method includes two steps, prior estimation and posterior inference.

The trend prior is estimated by the local slope sampling, which is similar with the implementation in Eq. \ref{eq: linear_reg}. Specifically, a curvature $L^2_k$ is sliced into a few curve segment using an offset-direction sliding window. The local slope of each curve segment is estimated by Eq. \ref{eq: linear_reg}, and the mean depth and offset of the curve segment also be recorded. After the slope estimation of $n_s$ sampling curve segment, we obtain a three-dimensional point set $\{(\bar{o}^{(k)}, \bar{d}^{(k)}, s^{(k)})\}_{k=1}^{n_s}$, where the three dimensions are the mean of offset, the mean of depth, and the local slope, respectively. Since the sampling location is discrete, we adapt a kernel estimation method to infer the whole slope field. The slope estimation at the location of $o$ offset and $d$ depth can be computed by: 
\begin{equation}\label{eq: prior_map}
M^s[o, d] \triangleq \frac{1}{C} \sum_{k=1}^{n_s} f_{\text{G}}((\bar{o}^{(k)}, \bar{d}^{(k)}; h^{\text{prior}}), (o, d)) * s^{(k)},
\end{equation}
where $C$ represents a partition function, and $h^{\text{prior}}$ indicates a bandwidth of $f_{\text{G}}(\cdot, \cdot; h)$, which is a Gaussian kernel function controlled by a bandwidth $h$:
\begin{equation}\label{eq: kernel_gauss}
f_{\text{G}}(\mathbf{x}_1, \mathbf{x}_2; h) \triangleq e^{-\frac{\|\mathbf{x}_1-\mathbf{x}_2\|^2_2}{2 h^2}},
\end{equation}
where $h$ indicates the bandwidth of Gaussian kernel. The estimated slope prior map $\mathbf{M}^s$ will provide the global trend of the slope for the local curvature regression. 

We conduct a posterior regression using the slope-constraint local linear regression, where each point of $\mathbf{L}^2_k$ will be regressed locally. 
Shortly, we only showcase the posterior inference of the point $(o^*, d^*)$ on the curvature $\mathbf{L}^2_k$. 
Based on the slope prior map $\mathbf{M}^s$, we assume that the prior distribution of the slope at the offset $o^*$ and the depth $d^*$ follows a Gaussian distribution with a mean of $M^s[o^*, d^*]$ and a variance of 1:
\begin{equation}\label{eq: gauss_slope}
p_{[o^*, d^*]}(s) = \mathcal{N}(s; M^s[o^*, d^*], 1).
\end{equation}

Subsequently, we derive the optimization function for prior constraint locally linear regression:
\begin{equation}\label{eq: kgauss_loss}
\begin{array}{rl}
  \mathcal{L}(o^*; b, s) = & \sum_{i=1}^{n^c} f_{\text{G}}(o^*, o_i; h_{\text{data}}) \left( d_i - \left( s\times o_i + b \right) \right)^2 \\
  \  & + \frac{\lambda}{2 h^2_{\text{para}}} (s - M^s[o^*, d])^2,
\end{array}
\end{equation}
where $h_{\text{data}}$ and $h_{\text{para}}$ are bandwidths of the kernel regression of data and the prior slope constraint, $\lambda$ is a balance weight, and $n^c$ represents the number of all points in the curvatures $\{\mathbf{L}^2_{k}\}_{k=1}^{K_2}$. For ease of calculation, we will derive matrix representations below. First, we define the optimized parameter $\theta \triangleq [b, s]^T$, the dependent variable $\mathbf{d} \triangleq [d_1, ..., d_n]^T$, and independent variables 
$$
\mathbf{X} \triangleq \left[\begin{array}{cccc}
  1 & 1 & ... & 1\\
  o_1 & o_2 & ... & o_n
\end{array}\right]^T.
$$
Then, the slope prior constraint can be represented by a matrix:
$$
\mathbf{W}^c \triangleq \text{diag}(0, M^s[o^*, d^*]^T),
$$
and the local regression weights computed by Gaussian kernel also can be indicated by a matrix:
$$
\mathbf{W}^k = \text{diag}(f_{\text{G}}(o^*, o_1), ..., f_{\text{G}}(o^*, o_n)).
$$
The normal equation can be derived from Eq. \ref{eq: kgauss_loss}:
\begin{equation}\label{eq: norm_eq}
\left( \mathbf{X}^T \mathbf{W}^k \mathbf{X} + \lambda \mathbf{W}^c \right) \theta = \mathbf{X}^T \mathbf{W}^k \mathbf{d} + \lambda M^s[o^*, d^*].
\end{equation}
The new regression prediction for the original $(o^*, d^*)$ is:
\begin{equation}\label{eq: solve_eq}
\hat{d} = \left( \mathbf{X}^T \mathbf{W}^k \mathbf{X} + \lambda \mathbf{W}^c \right)^{-1} \left( \mathbf{X}^T \mathbf{W}^k \mathbf{d} + \lambda M^s[o^*, d^*] \right)[1 \ o^*]^T.
\end{equation}
Consequently, the updated depth value of each point in $\{\mathbf{L}^2_{k}\}_{k=1}^{K_2}$ can be regressed using Eq. \ref{eq: solve_eq}, and we denote the updated curvatures as $\{\mathbf{L}^3_{k}\}_{k=1}^{K_3}$. 
Finally, to ensure the effectiveness of each curvature, we eliminate those curvatures with a length shorter than $n_{\text{min}}$.

\section{Generation of Synthetic Samples}
Under the supervised learning framework of RMO picking, a ticklish problem is lack of the manual annotation of RMO curvatures. To tackle this issue, this section introduces a randomly sampling method of the CIG generation. 

\subsection{Background Assumption}
\begin{figure}[!ht]
    \centering
    \includegraphics[width=3.2in]{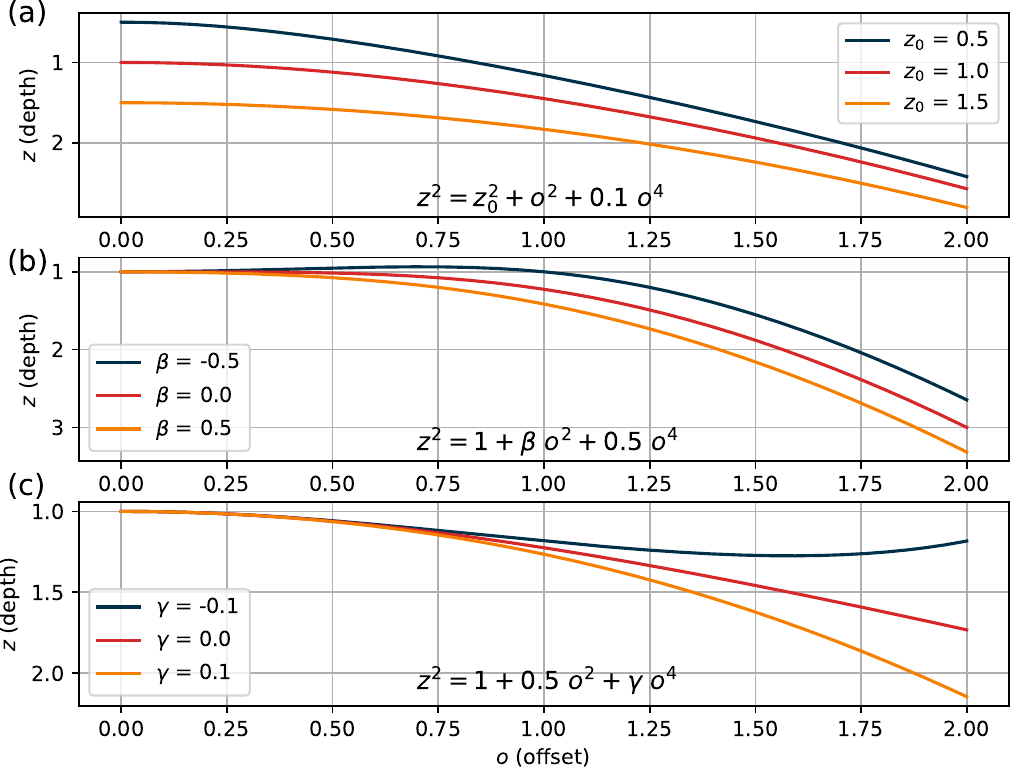}
    \caption{The influence of $z_0$, $\beta$ and $\gamma$ on the generated curvatures.}
    \label{fig: syn_function_show}
\end{figure}

Generally, residual curvatures in a CIG exhibit a form of hyperbolic equation. In this study, we further enhance the degrees of freedom of the equation:
\begin{equation}\label{eq: syn_eq}
z^2 = z_0^2 + \beta * o^2 + \gamma * o^4,
\end{equation}
where $z_0$ denotes the initial depth, and $o$ and $z$ represent two variables along the offset axis and the depth axis. Additionally, $\beta$ and $\gamma$ are two hyperparameters used to control the degree of wriggle. Fig. \ref{fig: syn_function_show} displays the influence of $z_0$, $\beta$ and $\gamma$ on the generated curvatures. Specifically, $z_0$ decides the zero-offset depth as shown in Fig. \ref{fig: syn_function_show}(b) and (c) illustrate the effects of $\beta$ and $\gamma$ on the degree of wriggling in the middle and right sections, respectively. In the following subsection, we will provide detailed information on the hyperparameter settings.

\subsection{CIG Generation}
Our synthetic CIG includes $K^c$ residual curvatures, where the depth of these curvatures is gradually increasing. Concretely, the start depth of $k$th curvature, denoted as $z_0^{(k)}$, is defined by a random initial depth $\epsilon^z \sim U(10, 200)$, in addition to a noisy depth that follows a ladder pattern: 
\begin{equation}\label{eq: syn_z0}
z_0^{(k)} = \epsilon^z + \left[\frac{d_{\max}}{K^c}+\epsilon^d\right]*k,
\end{equation}
where $d_{\max}$ is the maximum of depth, and $\epsilon^d$ represents a interval uniformly distributed sampling noise $ U(-0.0125, 0.0125)$.
Aiming to simulate the change under different depth, these curvatures are divided into four parts vertically. Each part is controlled by diverse $\beta$s and $\gamma$s, as shown in Tab. \ref{tab: syn_para}. However, the partial curvatures with far offset will intersect. Consequently, we employ a cropping technique to enhance the robustness of the generation process. Specifically, the point $(o_i, z_i)$ will be removed if
$$o_i > z_i * r^c + o_0$$
where $r^c$ is a cropping rate parameter, and $o_0$ represents the initial offset position.

\begin{table}[!ht]
    \centering
    \caption{Hyperparameter Setting of Synthetic CIG}
    \resizebox{\linewidth}{!}{
    \begin{tabular}{lcccc}
        \toprule
        \textbf{Curvature Index Range} & {$\mathbf{\beta}$} & $\mathbf{\gamma}$ \\ \hline
        $\left[1, \frac{1}{3}K^c\right)$ & $U(0.275,1.125)$ & $U(2.75\times 10^{-4},6.25\times 10^{-4})$ \\
        $\left[\frac{1}{3}K^c, \frac{1}{2}K^c\right)$ & $U(0.125,0.50)$ & $U(2.00\times 10^{-4},3.75\times 10^{-4})$ \\
        $\left[\frac{1}{2}K^c, \frac{2}{3}K^c\right)$ & $U(-0.50,0.125)$ & $U(-2.50\times 10^{-4},-1.25\times 10^{-4})$ \\ 
        $\left[\frac{2}{3}K^c, K^c\right]$ & $U(0.275,1.125)$ & $U(2.75\times 10^{-4},6.25\times 10^{-4})$ \\ 
        \bottomrule
    \end{tabular}}
    \label{tab: syn_para}
\end{table}

Finally, we conduct a convolution to generate the CIG, where a Ricker wavelet is used to simulate the waveform of CIG. The sampling numbers of the depth axis and the offset axis in a CIG are 1000 and 100, respectively. Concretely, the discrete implementation of convolution are defined by:
\begin{equation}\label{eq: ricker}
\begin{array}{l}
  h(z, z^c) = \left[\pi (z-z^c)f\right]^2, \\
  g[z, k] = \sum_{n=1}^{n^c_k}{\left[1 - 2h(z, z^c_{k, n})\right] e^{-h(z, z^c_{k, n})}},
\end{array}
\end{equation}
where $g[z, k]$ indicates the amplitude of the $z$th row and $k$th column of CIG, and $z^c_{k, n}$ represents the row index of $n$th curvature point on the $k$th column of CIG. Moreover, the frequency $f$ in Eq. \ref{eq: ricker} decreases from 16 Hz to 4 Hz with the depth of curvatures. Additionally, aiming to simulate acquisition noise, we also inject uniformly distributed noise with the power of $\pm 5\%$ original signal power.

A training set and a validation set are essential for a supervised learning task. Consequently, we generated 4,000 and 1,000 CIGs for training and validation, respectively. The training set consists of 1,000 CIGs, each with curvatures of 50, 60, 80, and 100. The validation set includes 500 CIGs, each with curvatures of 60 and 80. 

\section{New Metrics for RMO Picking}
To evaluate automatical RMO pickings, we define three metrics in terms of the tracking level and the picking trend. First, semblance offers a measure of the tracking level of each curvature picking without manual labels. The semblance value of $k$th curvature $\mathbf{L}_{k}$ is computed by:
\begin{equation}\label{eq: semblance}
s[k] = \frac{\sum_{m=-h_s}^{h_s}{\left(\sum_{i=1}^{n_k}{g[o_i, z_i+m]}\right)}^2}{n_k \sum_{m=-h_s}^{h_s} {\sum_{i=1}^{n_k}{g[o_i, z_i+m]^2}}},
\end{equation}
where $g$ represents a CIG, and $h_s$ indicates the length of the selected window. $s[k]$ provides a reference of the tracking level for single curvature. The global evaluation can be represented by the average of all semblance values. 

Second, track rate (TR) defines the rate at which automatic picking tracks manual picking. We assume the judgment standard of successful tracking for $k$th manual curvature $\mathbf{L}^M_{k}$, where if the average distance between any pick in the automatic pick collection and the current manual pick is less than the threshold $d^t$, it is considered a successful tracking versus $\mathbf{L}^M_{k}$. An indicative function can define the above judgment:
\begin{equation}\label{eq: tr_success}
I^t(\{\mathbf{L}^A_{k'}\}_{k'=1}^{K^A};\mathbf{L}^M_{k}) = I\left\{\exists \mathbf{L}^A_{k'} \in \{\mathbf{L}^A_{k'}\}_{k'=1}^{K^A}, d(\mathbf{L}^A_{k'}, \mathbf{L}^M_{k}) < d^t\right\},
\end{equation}
where $d(\cdot, \cdot)$ indicates the average values of the errors of the same offset for two curvatures, and $K^A$ is the number of automatic curvature pickings. Subsequently, TR can be defined by:
\begin{equation}\label{eq: tr}
\text{TR} = \frac{1}{K^M}\sum_{k=1}^{K^M}I^t(\{\mathbf{L}^A_{k'}\}_{k'=1}^{K^A};\mathbf{L}^M_{k}),
\end{equation}
where $K^M$ represents the number of manual curvature pickings. In this study, the threshold $d^t$ in Eq. \ref{eq: tr_success} is set to 3 pixels.

Third, mean square error (MSE) between manual and automatic slope fields measures the consistency of picking trend between automatic pickings and ground truth. Our proposed robust bayesian regression method offers a definition of the slope field (Eq. \ref{eq: prior_map}). For manual pickings $\{\mathbf{L}^M_{k}\}_{k=1}^{K^M}$ and automatic pickings $\{\mathbf{L}^A_{k}\}_{k=1}^{K^A}$, two slope field maps $M^{\text{slope}}_M$ and $M^{\text{slope}}_A$ can be calculated by Eq. \ref{eq: prior_map}, respectively. Subsequently, MSE between two maps is defined by:
\begin{equation}\label{eq: mse}
\text{MSE} = \frac{1}{n^{\text{offset}}n^{\text{depth}}}\sum_{i=1}^{n^{\text{offset}}}\sum_{j=1}^{n^{\text{depth}}}
{\left({M^{\text{slope}}_M}_{ij}-{M^{\text{slope}}_A}_{ij}\right)^2},
\end{equation}
where $n^{\text{offset}}$ and $n^{\text{depth}}$ represent sampling numbers of the offset domain and the depth domain in the slope map.

\section{Numerical Experiments}
In this section, our proposed cascade RMO picking method will be evaluated on both model data and field data. Additionally, we also compare the picking results of human and the semblance-based method with ours. 
\subsection{Implementation Details of Learning Process}
The IFSN in our proposed cascade RMO picking method is a neural network, whose weights must be learned prior to inference. The learning stage involves two iterative processes: training and validation, conducted on the synthetic datasets generated in Section III. To supervise the weight learning, we employ a masking strategy in which the pixels corresponding to the curvatures in the ground truth map are marked as 1, while all other pixels are labeled as 0. 

In the training process, several hyperparameters need to be tuned. In addition to the initial learning rate, batch size, and optimizer, we also consider an optimal combination of the loss function, the reduction factor of CBAM, and the optimization strategy, as illustrated in Tab. \ref{tab: hp-tuning}. To determine the optimal combination, we compute the Mean Intersection over Union (MIOU) metric between the segmentation map generated during the final validation process and the ground truth. The hyperparameter combination of the trained model that yields the minimum MIOU is regarded as the optimal choice within the parameter space outlined in Tab. \ref{tab: hp-tuning}. Specifically, to address the imbalance between positive and negative pixels, we attempted to utilize the focal loss function \cite{ross2017focal} during training. Unexpectedly, the BCE loss function performed better in the task of RMO picking. We analyzed that the focal loss is influenced by two hyperparameters, making it challenging to identify an optimal combination. Additionally, we employed a cosine optimization strategy, which adjusts the learning rate according to a cosine function. Experimental results indicate that the cosine optimization strategy can enhance learning quality. Furthermore, we implemented an early stopping training strategy, which halts training if the validation loss does not decrease for five consecutive epochs, with a maximum epoch limit set to 50 in our study.

\begin{table}[!ht]
    \centering
    \caption{Hyperparameter Setting of Training Processing}
    \resizebox{\linewidth}{!}{
    \begin{tabular}{lcc}
    \toprule
    \textbf{Hyperparameter Name}  & \textbf{Selection Set}       & \textbf{Optimal Parameter} \\ \hline
    Loss function            & BCE, Focal Loss\cite{ross2017focal}               & BCE     \\
    Reduction factor in CBAM & 8, 16, 32                    & 16      \\
    Initial learning rate    & 1e-2, 1e-3, 2e-3, 5e-3, 1e-4 & 1e-4    \\
    Batchsize of training    & 8, 16, 32                    & 8       \\ 
    Optimizer                & SGD, Adam, AdamW\cite{Loshchilov2017DecoupledWD}             & SGD     \\ 
    Optimization Strategy    & Constant, Cosine             & Cosine  \\ 
    \bottomrule
    \end{tabular}}
    \label{tab: hp-tuning}
\end{table}

\subsection{Test on Model Data and Field Data} 
\subsubsection{Introduction of Used Datasets}
\begin{figure}[!ht]
    \centering
    \subfloat[Synthetic]{\includegraphics[height=2.8in]{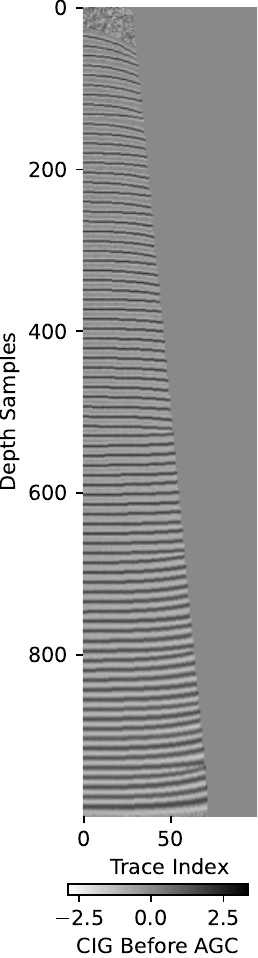}\label{fig: cig_syn}}
    \subfloat[BP]{\includegraphics[height=2.8in]{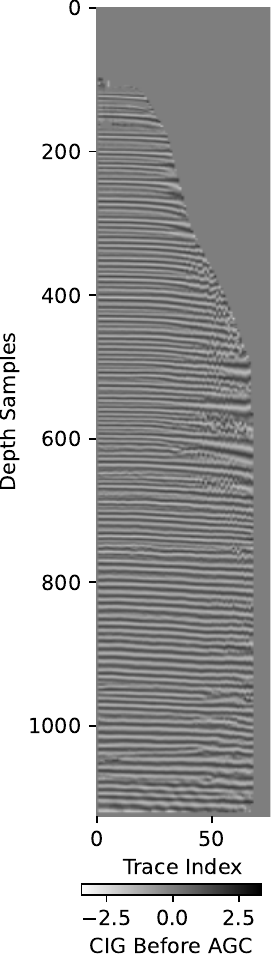}\label{fig: cig_bp}}
    \subfloat[F-A]{\includegraphics[height=2.8in]{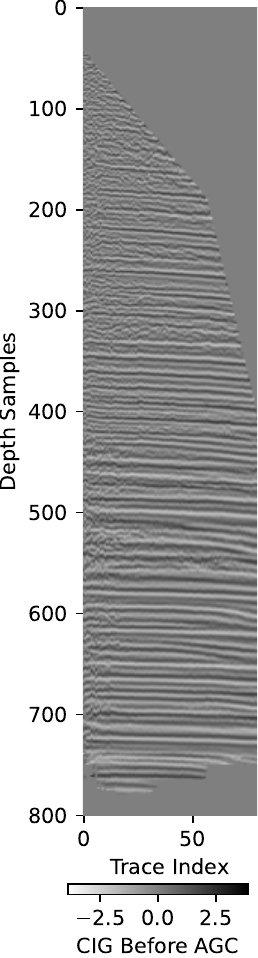}\label{fig: cig_sh}}
    \subfloat[F-B]{\includegraphics[height=2.8in]{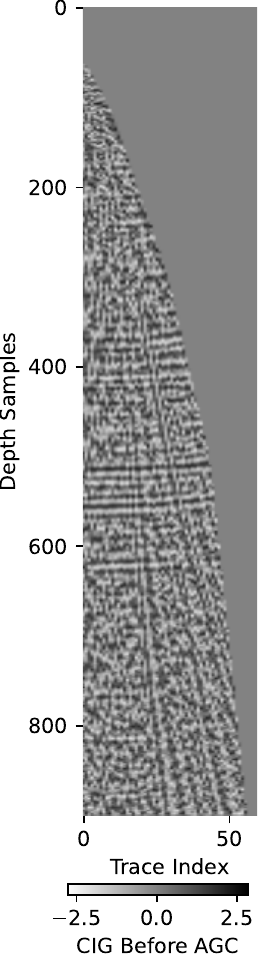}\label{fig: cig_yz}}
    \caption{Classic CIG Image of Synthetic, BP, F-A, and F-B, respectively.}
    \label{fig: data_visual}
\end{figure}

The synthetic CIGs generated by Section III can simulate the residual curvatures as shown in Fig. \ref{fig: cig_syn}, but the complex noises, e.g., multiple noise and angle noise, are not mimicked. 
Consequently, there is an obvious difference between the distribution of synthetic data and field data. 
Aiming to evaluate the practicability of our proposed cascade method in case of various SNRs, we select one open-source model dataset and two field datasets provided by China National Petroleum Corp Bureau of Geophysical Prospecting Inc. 
First, a model dataset named BP is used in our study, which was created by Hemang Shah and provided courtesy of BP Exploration Operation Company Limited. As shown in Fig. \ref{fig: cig_bp}, compared with synthetic CIGs, the CIGs of BP includes more complex low-frequency noise at the far-offset region. Second, we choose a field dataset with high SNR denoted F-A, as shown in Fig. \ref{fig: cig_sh}. Although F-A dataset has a higher SNR, the waveform frequency of the shallow residual curvature is significantly lower than that of the synthesized data, and there is more acquisition noise at the near offset, which brings challenges to the picking algorithm. To further evaluate the picking performance in case of low SNR, we finally select a field dataset with both multiple noise and angle noise named F-B, as shown in Fig. \ref{fig: cig_yz}. The residual curvatures in this dataset is aliased in noise, and the number of valid, clear curvatures is significantly lower than in both previous datasets. For each dataset, experts of seismic data processing provide 20 samples with the annotation of RMOs. More details of each dataset is illustrated in Tab. \ref{tab: dataset_intro}.

\begin{table}[!ht]
    \centering
    \caption{Basic Information of Used Datasets}
    \resizebox{0.9\linewidth}{!}{
    \begin{tabular}{lccc}
    \toprule
    \textbf{Site Name}  & \textbf{CIG Number} & \textbf{Use}         & \textbf{Labeled Number}\\ \hline
    Synthetic           & 6000                & Train \& Validation  & 6000                   \\
    BP                  & 1791                & Test                 & 20                   \\
    F-A                 & 3006                & Test                 & 20                   \\
    F-B                 & 15301               & Test                 & 20                   \\
    \bottomrule
    \end{tabular}}
    \label{tab: dataset_intro}
\end{table}

\subsubsection{Inference Hyperparameters}
In the inference process, our proposed cascading method has multiple hyperparameters, namely, the window length of AGC ($h$ in Eq. \ref{eq: AGC-E}), the two clustering hyperparameters of DBSCAN ($n^{\text{min}}$ and $d^{\text{eps}}$), the three bandwidth parameters in the Bayesian method ($h^{\text{prior}}$ in Eq. \ref{eq: prior_map}, and $h^{\text{data}}$ and $h^{\text{para}}$ in Eq. \ref{eq: kgauss_loss}), and the minimum length of picked curvature ($n_{\text{min}}$).
Since the data quality and the interval of residual curvatures differ among the BP, F-A, and F-B datasets, we must establish distinct inference hyperparameters for each dataset.
Tab. \ref{tab: infer_para} illustrates the detailed configuration of inference hyperparameters. Specifically, the parameters of AGC significantly influence the quality of the input feature map, allowing for a tailored set of parameters to be established for each dataset based on expert experience. In our experiment, the amplitude levels of the BP dataset vary dramatically at different depths; therefore, we opted for a smaller AGC window to mitigate these abrupt changes.
Additionally, the BP and F-A datasets exhibit a high SNR, and we tend to concatenate more curvatures; therefore, the parameter settings for DBSCAN favor the combination of multiple curvatures. Furthermore, the Bayesian bandwidth parameter is relatively insensitive, so we set it to a consistent value for each dataset. $n_{\text{min}}$ restricts the length of the final output curvature, so we set it to 20 for datasets with a high SNR where the effective curvature is complete, and 10 for the low SNR dataset where some curvature is separated by noise. 

\begin{table}[!ht]
    \centering
    \caption{Inference Hyperparameters}
    \resizebox{0.9\linewidth}{!}{
    \begin{tabular}{lcccc}
    \toprule
    \textbf{Site Name}  & $(h_1, h_2)$ & $(n^{\text{min}}, d^{\text{eps}})$ & $(h^{\text{prior}}, h^{\text{data}}, h^{\text{para}})$ & $n_{\text{min}}$  \\ \hline
    BP                  & (9, 15)  & (1, 8)   & (5, 5, 50) & 20 \\
    F-A                 & (15, 31) & (1, 8)   & (5, 5, 50) & 20 \\
    F-B                 & (15, 31) & (1, 4)   & (5, 5, 50) & 10 \\
    \bottomrule
    \end{tabular}}
    \label{tab: infer_para}
\end{table}

\begin{figure*}[!ht]
    \centering
    \subfloat[CIG]{\includegraphics[height=3.4in]{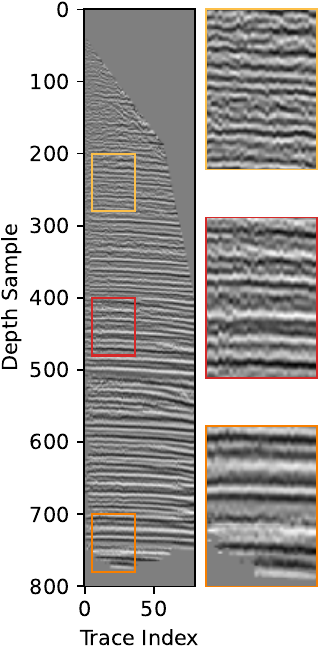}\label{fig: cp_SH_1}}
    \subfloat[Segmentation]{\includegraphics[height=3.4in]{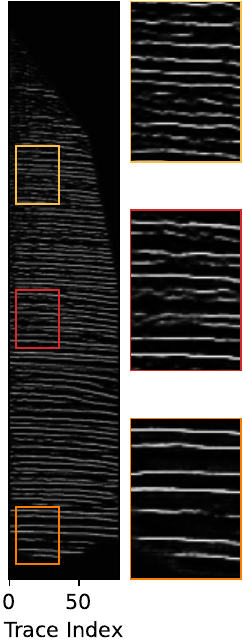}\label{fig: cp_SH_2}}
    \subfloat[Raw RMOs after CG]{\includegraphics[height=3.4in]{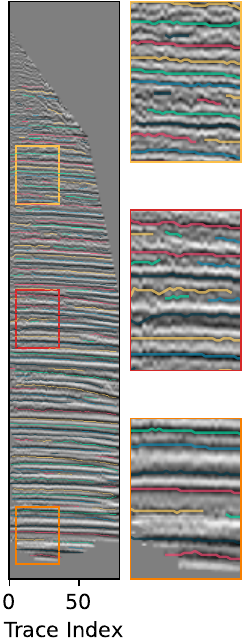}\label{fig: cp_SH_3}}
    \subfloat[Clustering]{\includegraphics[height=3.4in]{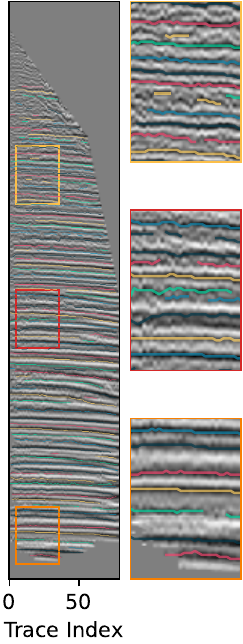}\label{fig: cp_SH_4}}
    \subfloat[Bayesian Regression]{\includegraphics[height=3.4in]{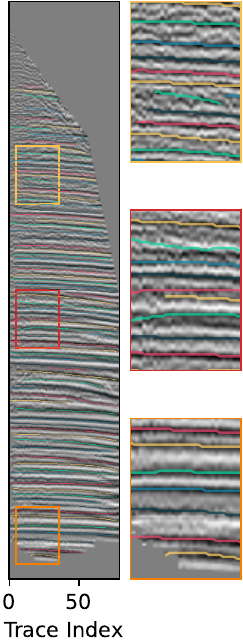}\label{fig: cp_SH_5}}
    \caption{Detailed output of each step in inference process of the cascade method. (a) The CIG after AGC with the window of 15; (b)-(d) The output results of Step 1-4 of the cascade method.}
    \label{fig: cp_picking_process}
\end{figure*}

\subsubsection{Picking Performance of Cascade Method}
Initially, we present the detailed outputs of each step in our cascade method to illustrate the impact of each component. To avoid repetitive descriptions, we will use a CIG of the F-A dataset as an example. The CIG image following the AGC operation is visualized in Fig. \ref{fig: cp_SH_1} The overall trend of the RMOs is horizontal; however, there is local jitter that cannot be accurately described by polynomial equations. The segmentation map inferred by the IFSN is presented in Fig. \ref{fig: cp_SH_2}, where the local texture of RMOs can be captured in this field CIG. This demonstrates that the IFSN, trained on synthetic data, can be effectively transferred to predict outcomes in a field domain. Fig. \ref{fig: cp_SH_3}-\ref{fig: cp_SH_5} illustrate the output curvatures for Steps 2, 3, and 4, respectively. First, raw curvatures are extracted using the CG method, which results in picked curvatures that are discontinuous and exhibit high-frequency jitter. To address the issue of discontinuity, we propose a density clustering method based on our new definition of distance, which merges curvatures belonging to a single RMO. The merged curvatures are illustrated in Fig. \ref{fig: cp_SH_4}. Subsequently, our proposed Bayesian regression method smooths the current RMOs to ensure alignment with the local slope trend. Fig. \ref{fig: cp_SH_5} demonstrates that the Bayesian regression method not only connects curvatures within the same cluster but also preserves the approximate picking trend of the regression curves.

\begin{figure*}[!ht]
    \centering
    \subfloat[Segmentation (BP)]{\includegraphics[height=1.5in]{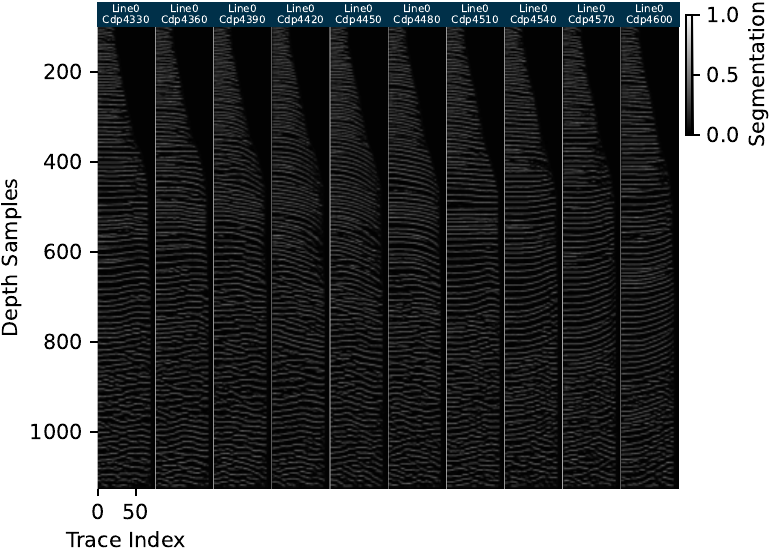}\label{fig: cp_BP_seg}}
    \subfloat[Segmentation (F-A)]{\includegraphics[height=1.5in]{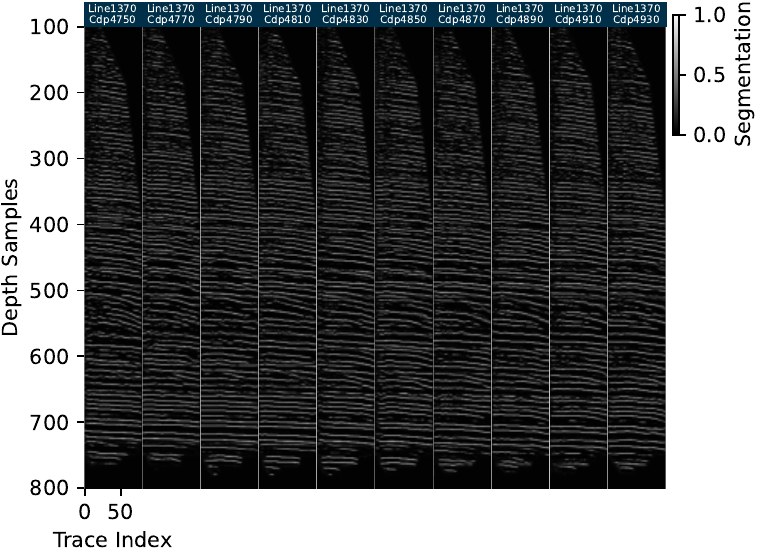}\label{fig: cp_SH_seg}}
    \subfloat[Segmentation (F-B)]{\includegraphics[height=1.5in]{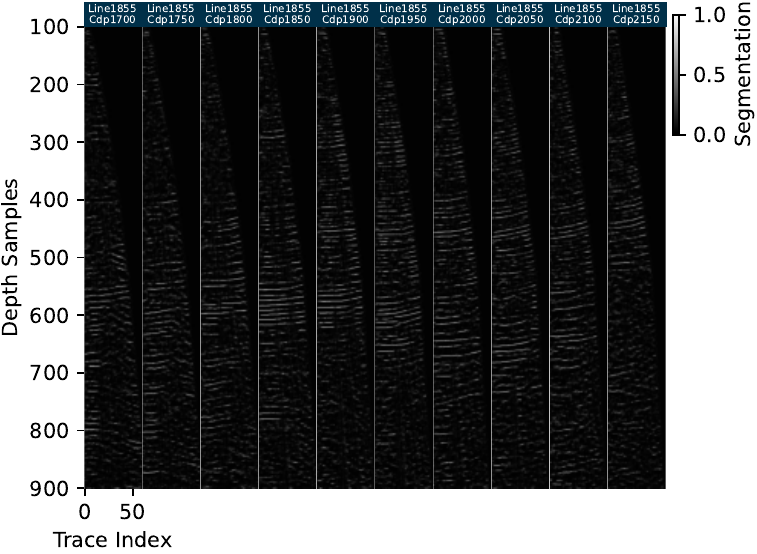}\label{fig: cp_YZ_seg}}\\
    \subfloat[RMO Picks (BP, Cascade)]{\includegraphics[height=1.5in]{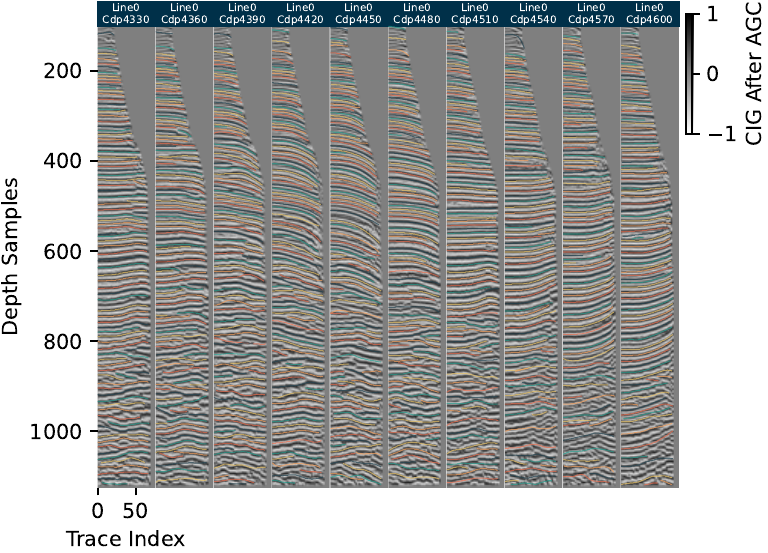}\label{fig: cp_BP_picks}}
    \subfloat[RMO Picks (F-A, Cascade)]{\includegraphics[height=1.5in]{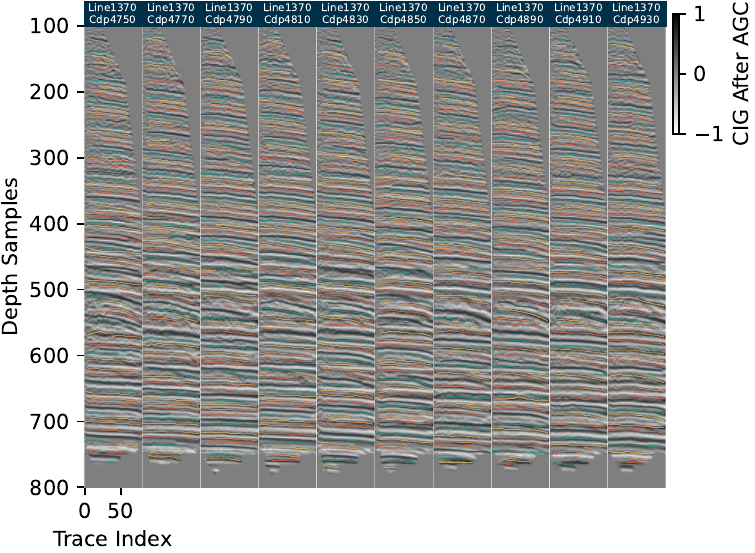}\label{fig: cp_SH_picks}}
    \subfloat[RMO Picks (F-B, Cascade)]{\includegraphics[height=1.5in]{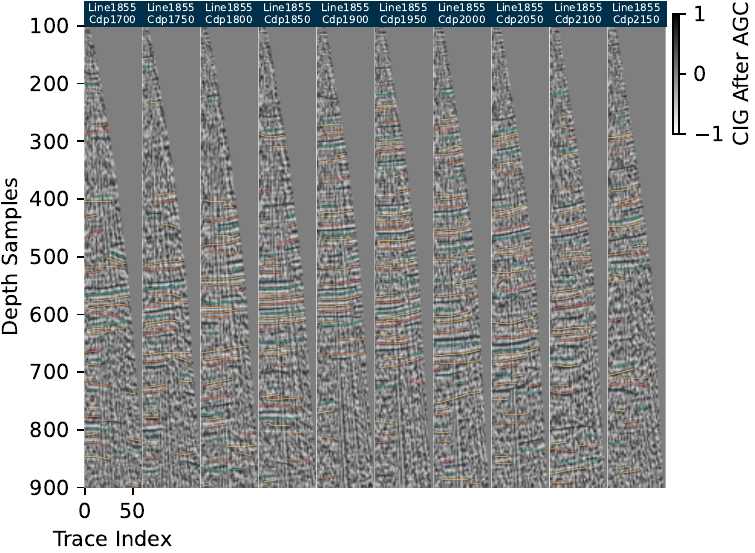}\label{fig: cp_YZ_picks}}\\
    \subfloat[Slope Field (BP, Cascade)]{\includegraphics[height=1.5in]{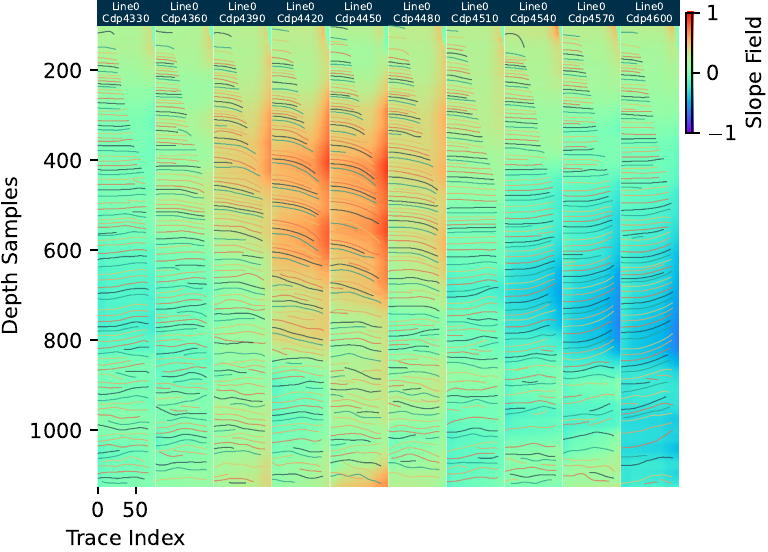}\label{fig: cp_BP_slope}}
    \subfloat[Slope Field (F-A, Cascade)]{\includegraphics[height=1.5in]{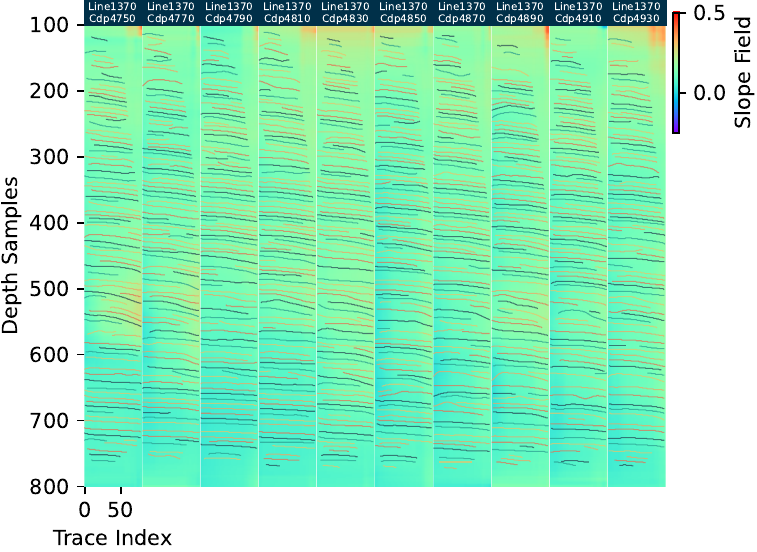}\label{fig: cp_SH_slope}}
    \subfloat[Slope Field (F-B, Cascade)]{\includegraphics[height=1.5in]{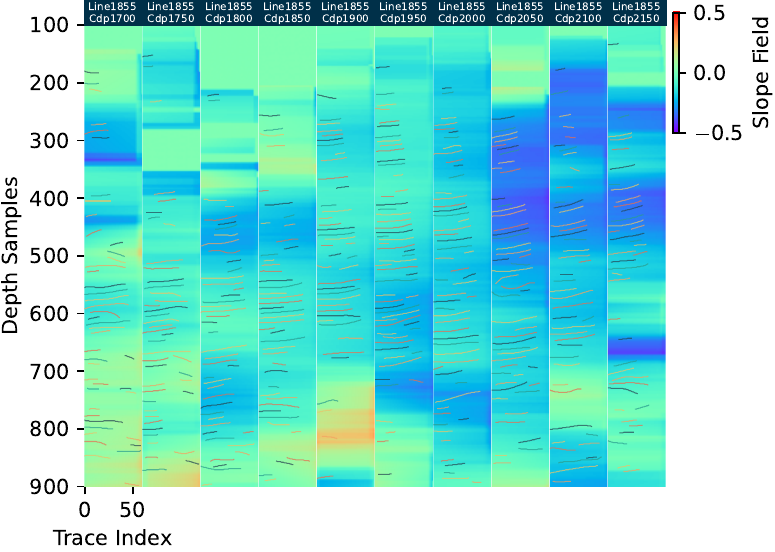}\label{fig: cp_YZ_slope}}\\
    \subfloat[RMO Picks (BP, Semblance)]{\includegraphics[height=1.5in]{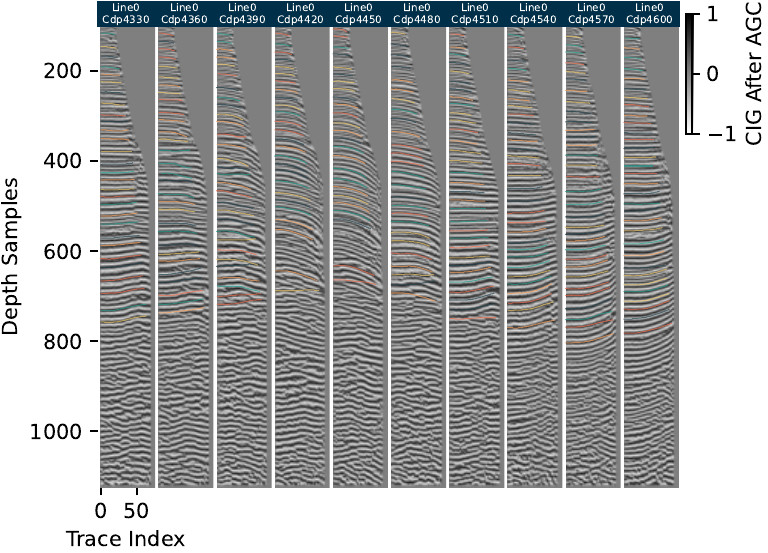}\label{fig: sp_BP_picks}}
    \subfloat[RMO Picks (F-A, Semblance)]{\includegraphics[height=1.5in]{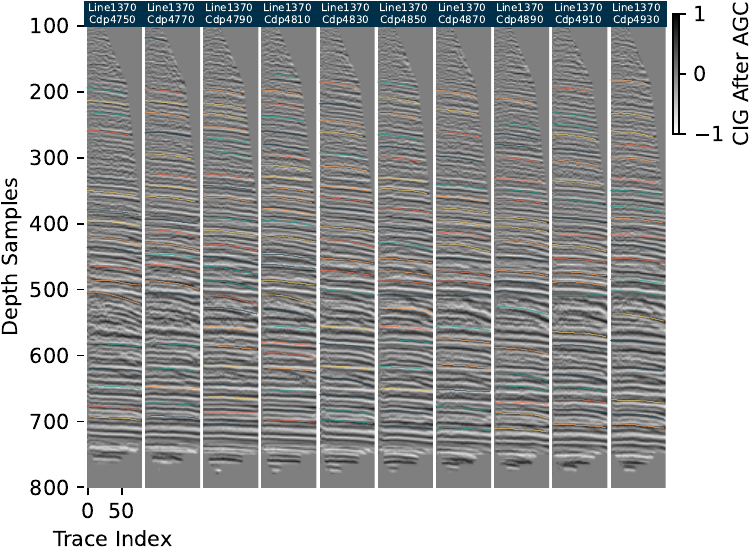}\label{fig: sp_SH_picks}}
    \subfloat[RMO Picks (F-B, Semblance)]{\includegraphics[height=1.5in]{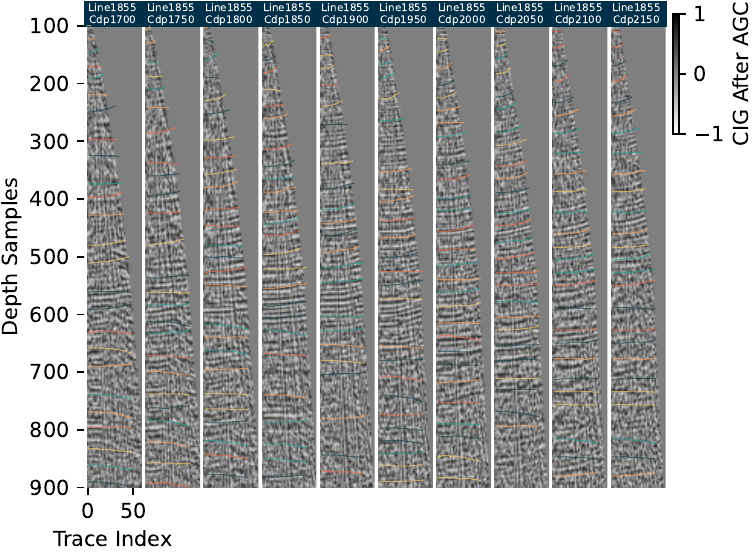}\label{fig: sp_YZ_picks}}\\
    \subfloat[Slope Field (BP, Semblance)]{\includegraphics[height=1.5in]{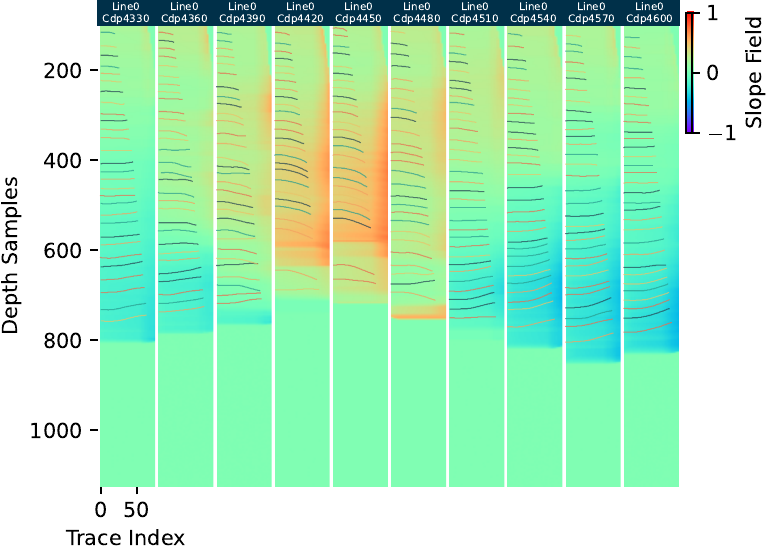}\label{fig: sp_BP_slope}}
    \subfloat[Slope Field (F-A, Semblance)]{\includegraphics[height=1.5in]{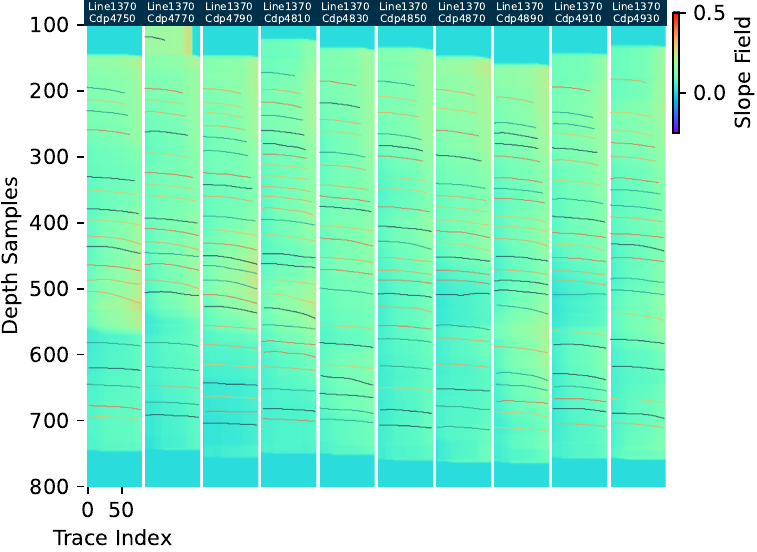}\label{fig: sp_SH_slope}}
    \subfloat[Slope Field (F-B, Semblance)]{\includegraphics[height=1.5in]{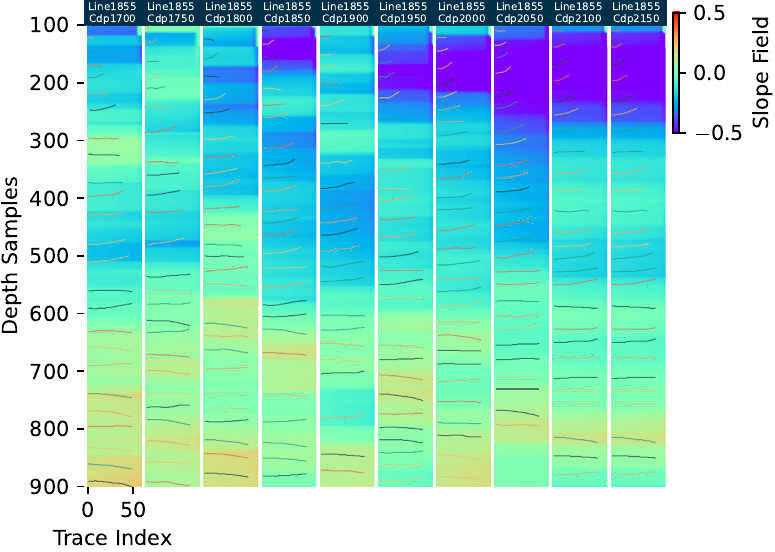}\label{fig: sp_YZ_slope}}
    \caption{Line prediction of the cascade method and the semblance method.}
    \label{fig: performance_cp}
\end{figure*}

To further evaluate the global accuracy of the cascade method, we visualize the line prediction of each dataset, as shown in Fig. \ref{fig: performance_cp}. 
Fig. \ref{fig: cp_BP_picks} - \ref{fig: cp_BP_slope} present the picking results for the BP dataset. Since the single in this dataset is relatively pure, the picked RMOs are continuous from left to right as shown in Fig. \ref{fig: cp_BP_picks}. Particularly, the deep-layer region includes various complex curvatures, which is not described by a equation with an order lower than three. The IFSN in our cascade method achieves capturing each local variation with any assumption of the curvature equation, as shown in Fig. \ref{fig: cp_BP_seg}. Subsequently, the images of the slope field for the final picks (Fig. \ref{fig: cp_BP_slope}) also demonstrate the robustness of picks for the BP dataset, where the slope values of the adjacent (Common Depth Point) CDP in Fig. \ref{fig: cp_BP_slope} are smooth. 

Fig. \ref{fig: cp_SH_picks}-\ref{fig: cp_YZ_slope} present the picks for the field datasets (F-A and F-B). In the picking of the field data with high SNR (F-A), the automatic RMO picks are flatten because of the high-quality initial velocity. Fig. \ref{fig: cp_SH_seg} visualizes the segmentation of IFSN for the F-A dataset. Although the vertical density of the RMOs is higher than the learned synthetic CIGs (Fig. \ref{fig: cig_syn}), each RMO can be recognized at the step of segmentation. Fig. \ref{fig: cp_SH_picks} also evaluates that our cascade method can split two closed curvatures through our proposed post-processing method, in which it is difficult for the single clustering method-based post-processing \cite{Bazargani2022segmentation} to divide them into two RMOs from the segmentation map (Fig. \ref{fig: cp_SH_seg}). This proves the practicability of our method in high density curvature picking.

\begin{figure*}[!h]
    \centering
    \subfloat[Case 1 (BP)]{\includegraphics[height=2.4in]{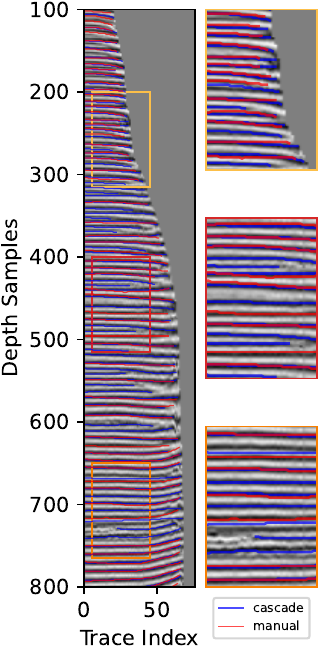}\label{fig: mp_BP_1}}
    \subfloat[Case 2 (BP)]{\includegraphics[height=2.4in]{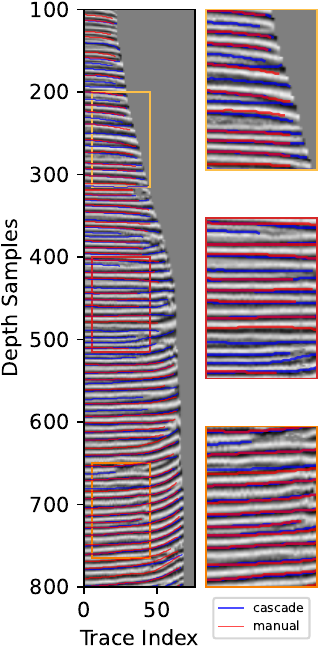}\label{fig: mp_BP_2}}
    \subfloat[Case 1 (F-A)]{\includegraphics[height=2.4in]{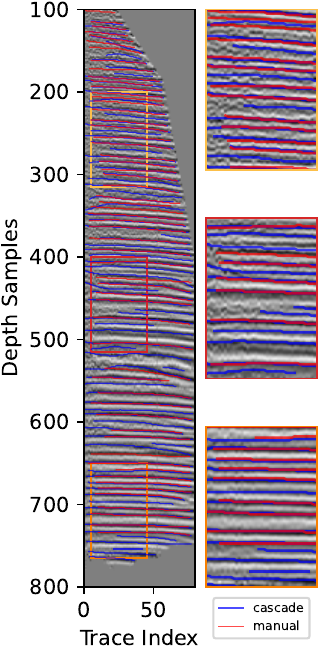}\label{fig: mp_SH_1}}
    \subfloat[Case 2 (F-A)]{\includegraphics[height=2.4in]{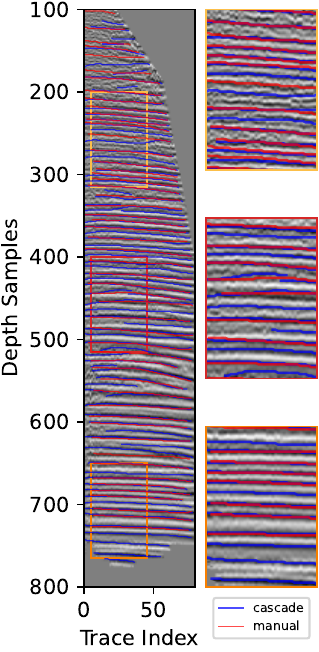}\label{fig: mp_SH_2}}
    \subfloat[Case 1 (F-B)]{\includegraphics[height=2.4in]{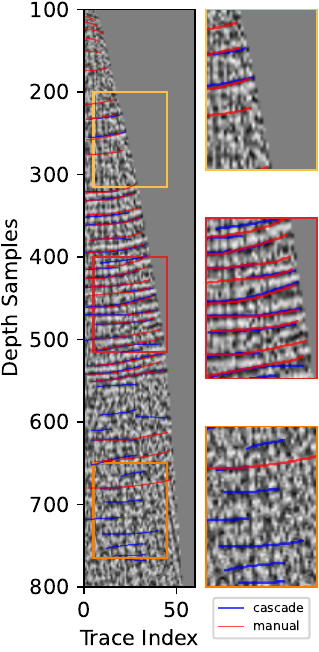}\label{fig: mp_YZ_1}}
    \subfloat[Case 2 (F-B)]{\includegraphics[height=2.4in]{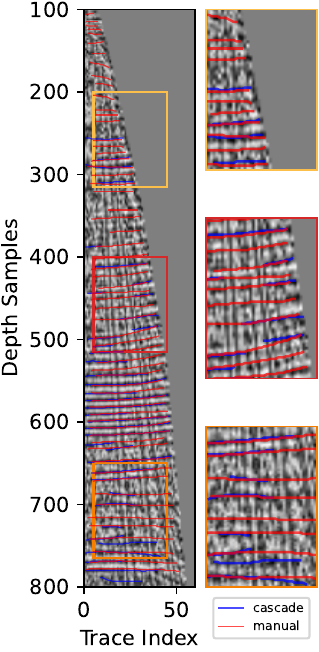}\label{fig: mp_YZ_2}}\\
    \subfloat[Slope Field (BP)]{\includegraphics[height=1.4in]{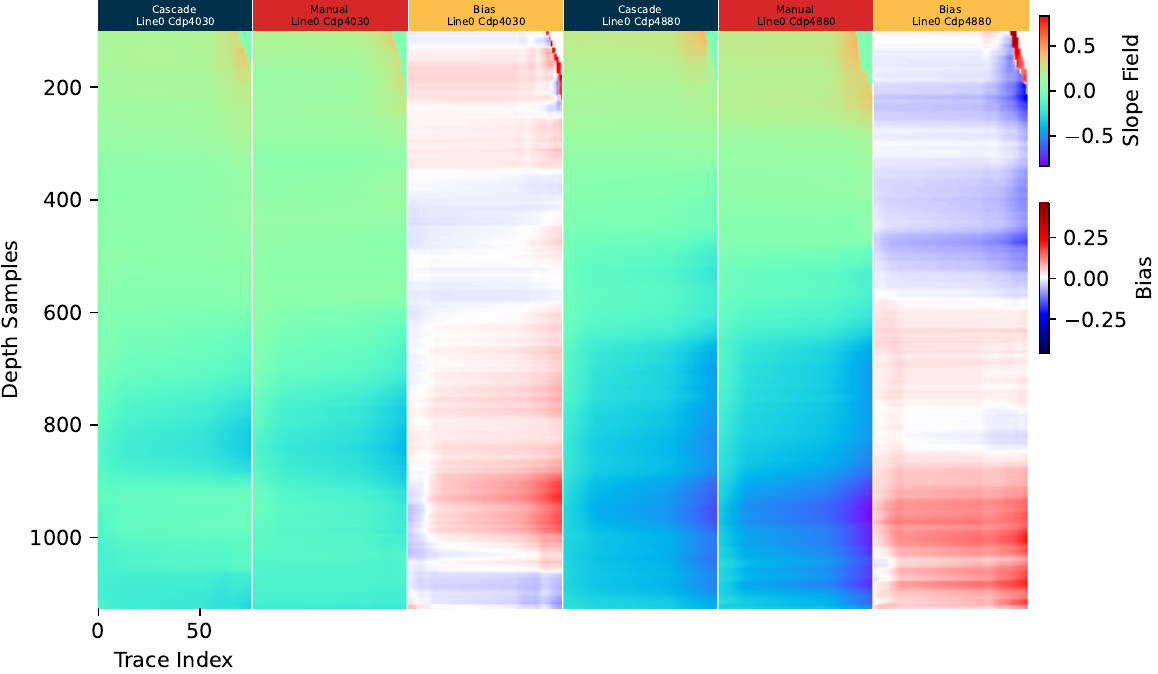}\label{fig: mp_BP_slope}}
    \subfloat[Slope Field (F-A)]{\includegraphics[height=1.4in]{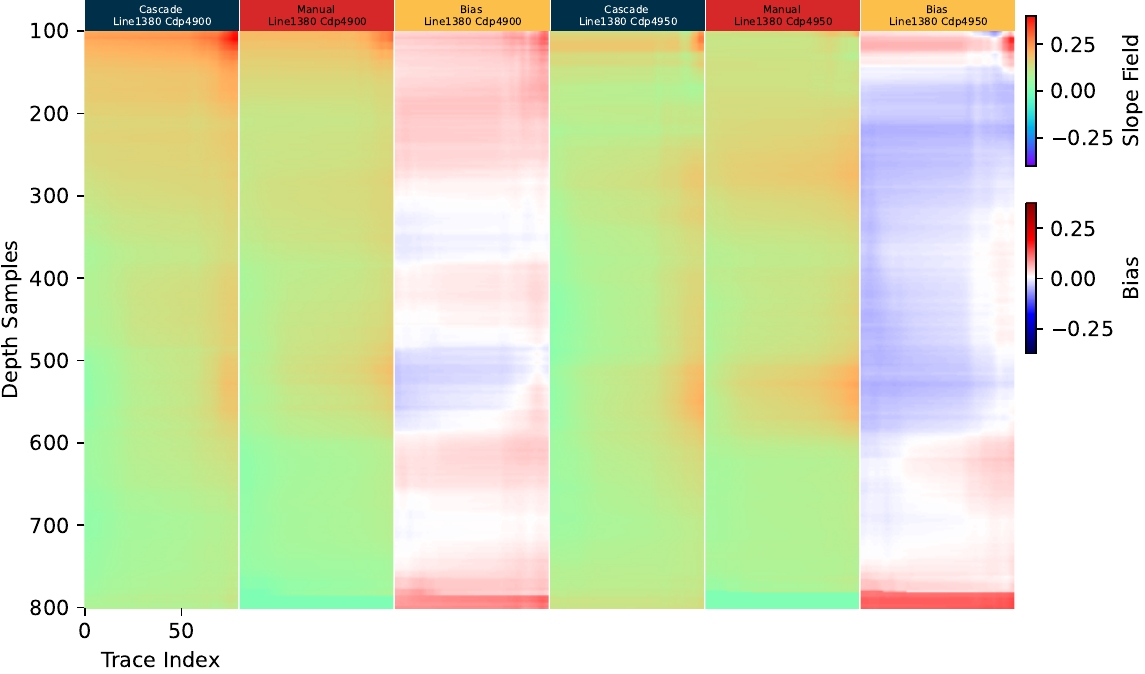}\label{fig: mp_SH_slope}}
    \subfloat[Slope Field (F-B)]{\includegraphics[height=1.4in]{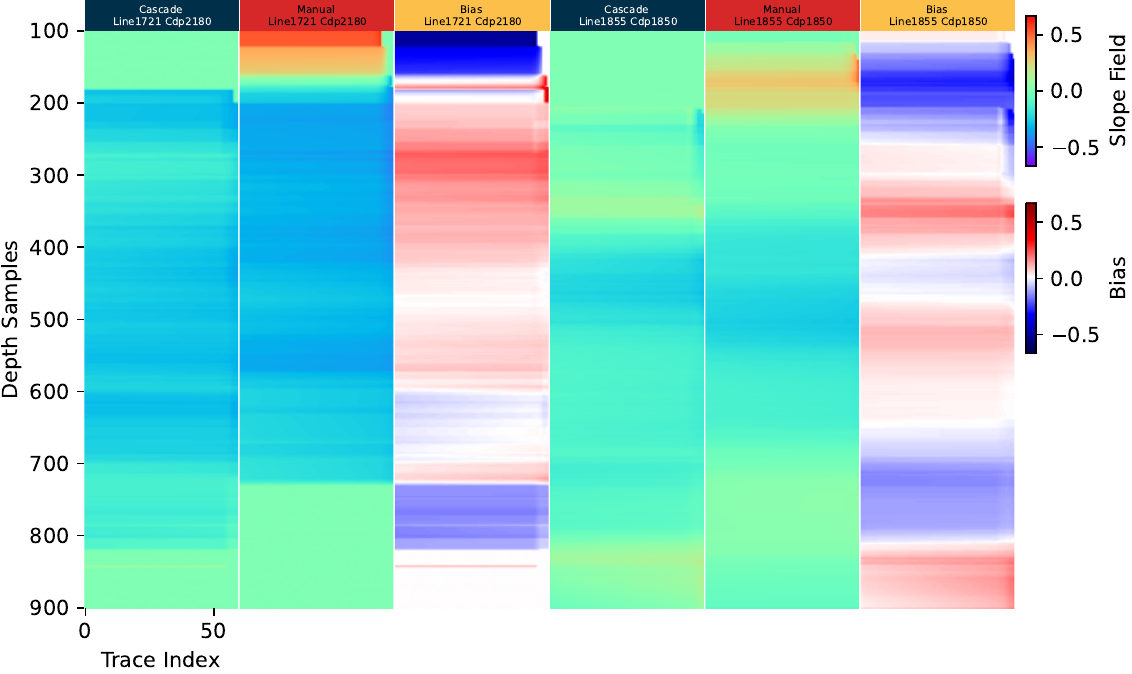}\label{fig: mp_YZ_slope}}
    \caption{Comparison of the cascade method and manual picks.}
    \label{fig: comp_w_mp}
\end{figure*}

In real picking scenarios, datasets often do not have the same high SNR as the BP and F-A datasets. Consequently, it is most important to avoid noise interference and pick effective curvatures at low SNR. Fig. \ref{fig: cp_YZ_picks} visualizes the picks of the cascade method for the field CIG with low SNR (F-B). Compared with the valid curvatures in the BP and F-A datasets, its number of F-B is extremely low. Thus, the cascade only capture up more RMOs in the middle layer, and less RMOs in the shallow and deep layers. It causes the abrupt variation of slope values at the shallow and deep layers, as indicated in Fig. \ref{fig: cp_YZ_slope}. However, we believe that the collected RMOs is accurate and can provide more residual information for the tomography inversion. Through this experiment, we also verify that the model trained on the synthetic data can identify the CIG containing a small amount of RMO and abundant complex noises, which has certain reference significance for other tasks in the field of geophysics without manual labels. 

\subsubsection{Compared with Semblance-based Method}
Currently, the semblance-based picking method is the most popular automatic picking method in the field application. In this study, a semblance-based method with a cubic polynomial assumption is selected. The RMO picks of the semblance method are illustrated in Fig. \ref{fig: sp_BP_picks}-\ref{fig: sp_YZ_slope}. In the picking process of the semblance method, the quality factor calculated in advance determines the quantity and starting position of the RMO pick. It causes no pick at the deep regions of BP (Fig. \ref{fig: sp_BP_slope}) and the sparse picks on both the F-A (Fig. \ref{fig: sp_SH_slope}) and the F-B (Fig. \ref{fig: sp_YZ_slope}). 

In term of the numbers of picked RMOs, our proposed cascade method surpasses the semblance method greatly. 
In terms of tracking details, the cascade method can track RMOs that cannot be expressed by cubic polynomials compared with the semblance method, which also reflects the improvement in the accuracy of our method in the task of RMO picking. 
For the overall trend of picking, the cascading method, while being able to depict more detailed changes, can maintain the local similarity of the trend as well as the semblance method, with the comparison between Fig. \ref{fig: cp_BP_slope} and Fig. \ref{fig: sp_BP_slope} being the most obvious.

Certainly, the semblance method also has its advantages. Specifically, as long as the quality factor requires a pick at this depth, the corresponding quadratic curve will eventually be picked. However, this advantage also comes with certain drawbacks. For instance, when the SNR at this depth is not suitable for picking, e.g., the depth of 100-300 and 700-900 in Fig. \ref{fig: sp_YZ_picks}, the picked curvatures often does not correspond to the true RMOs, resulting in an invalid pick. Based on the aforementioned visual comparisons, we have concluded that, in comparison to the semblance method, our approach is capable of delineating RMOs with higher precision while maintaining the same picking trend. The comparison of quantitative results will be discussed in detail in the following subsection.

\subsubsection{Compared with Manual Picking}
Aiming to evaluate the accuracy of the cascade method, we compare our picks with manual picking. As indicated in Table \ref{tab: dataset_intro}, experts label 20 samples for each dataset, where the annotated RMOs are considered the optimal choice when only CIGs are observed.

Fig. \ref{fig: comp_w_mp} illustrates the picking results for two classic cases from each dataset. Overall, Fig. \ref{fig: mp_BP_slope} and \ref{fig: mp_SH_slope} demonstrate that the cascade method and manual picking exhibit a consistent picking trend. Specifically, the detailed comparison presented in Fig. \ref{fig: mp_BP_1}-\ref{fig: mp_SH_2} show that picking density of our picks is higher than that of the manual picks. This discrepancy arises because manual picking is limited by workload constraints, whereas the IFSN can identify each RMO effectively. Consequently, this intensive picking in high-precision tomography inversion can yield more accurate RMO information, leading to a more efficient inversion iteration.
In the presence of a low SNR, Fig. \ref{fig: mp_YZ_1} and \ref{fig: mp_YZ_2} demonstrate that the cascade method is capable of recognizing the RMOs under the severe angle noise. Even in the case of weak RMO signals, a small amount of effective RMO can still be extracted. The limited number of shallow and deep picks contributes to the close alignment of our mid-layer picks with those made by human analysts, as illustrated in Fig. \ref{fig: mp_YZ_slope}.

Aiming for an objective comparison, we calculate the metrics for the picks made by both the cascade method and the semblance-based method, utilizing three metrics proposed in Section IV. Tab. \ref{tab: comp_test} presents the detailed test results. Since semblance only indicates the degree of coherence, it does not provide an accurate measure of absolute picking precision. Consequently, the corresponding index for manual picking will be lower than that for automatic picking.
TR reflects the recognition rate of effective curvature. The results indicate that the semblance-based method has a detection rate of less than 50\% for effective signals across all three datasets. In contrast, the cascade method achieves a picking rate of 90\% on both the model and field datasets with high SNR, and even reaches a picking rate of 61.31\% on the F-B dataset, which has a lower SNR. Furthermore, MSE of the slope field effectively measures the trend differences between automatic and manual picking. The findings suggest that, across all three datasets, our cascade method aligns more closely with the trend of manual picking compared to the semblance method. Based on these results, we believe that our cascade method not only surpasses the commonly used semblance method but also meets industrial requirements.

\begin{table}[!ht]
    \centering
    \caption{Comparison Test}
    \resizebox{\linewidth}{!}{
    \begin{tabular}{llccc}
    \toprule 
    \textbf{Site Name}  & \textbf{Method} & \textbf{Semblance} $\uparrow$ & \textbf{TR} $\uparrow$ & \textbf{MSE of Slope Field} $\downarrow$\\ \hline
    BP                  & Semblance       & 0.7611             & 0.4365      & 0.0639                     \\
                        & Cascade         & \textbf{0.7624}    & \textbf{0.9116} & \textbf{0.0058}                     \\
                        & Manual          & 0.6941             & -          &    -                      \\
    F-A                 & Semblance       & \textbf{0.6309}    & 0.3351      & 0.0138                     \\
                        & Cascade         & 0.6027             & \textbf{0.9153}      & \textbf{0.0017}                     \\
                        & Manual          & 0.5266             &  -          &    -                        \\
    F-B                 & Semblance       & 0.1015             & 0.2853      & 0.0382                     \\
                        & Cascade         & \textbf{0.3642}    & \textbf{0.6131}      & \textbf{0.0159}                     \\
                        & Manual          & 0.2598             &  -         &     -                     \\
    \bottomrule
    \end{tabular}}
    \label{tab: comp_test}
\end{table}

Based on the comparison with manual picking, we verify that our picking method can pick RMOs with high density in CIG with medium to high SNR, and the picking trend is consistent with manual picking. In low-SNR CIG, complex noise interference can be mitigated and effective RMO can be identified.

\section{Discussion}
\subsection{Ablation Study for IFSN}
IFSN, a segmentation network, is developed from U-Net \cite{ronneberger2015u}. By incorporating an attention mechanism, adjusting the activation function, and adding feature inputs, we have tailored IFSN for the task of curvature recognition.

To effectively retain both the positive and negative polarity features in the seismic signal, we replace the activation function in the first layer with TanH instead of ReLU. Additionally, aiming to efficiently integrate information from multiple feature maps, we add CBAM attention module after the first down-sampling module. Furthermore, to enhance the network's ability to perceive curvature features, we introduce a band-pass filter feature map ($F_{\text{BP}}$) and a peak feature map ($F_{\text{Peak}}$). In the following sections, we will systematically remove each module, retrain the model, and evaluate its performance on the F-B dataset to assess its effectiveness in practical applications.

\begin{figure}[!h]
    \centering
    \subfloat[Ablation Table]{\includegraphics[height=1.5in]{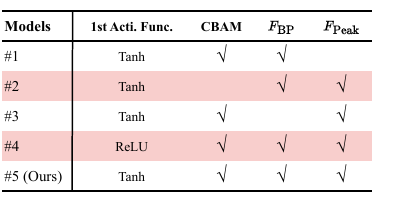}\label{fig: abl_tab}}\\
    \subfloat[Semblance $\uparrow$]{\includegraphics[height=0.9in]{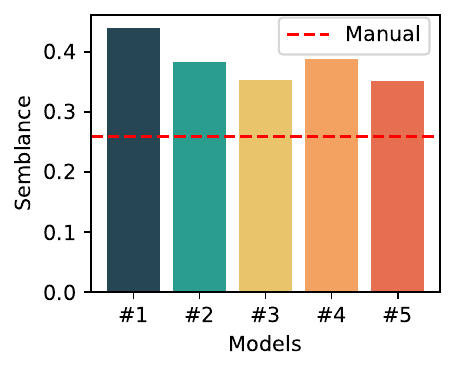}\label{fig: abl_sem}}
    \subfloat[TR $\uparrow$]{\includegraphics[height=0.9in]{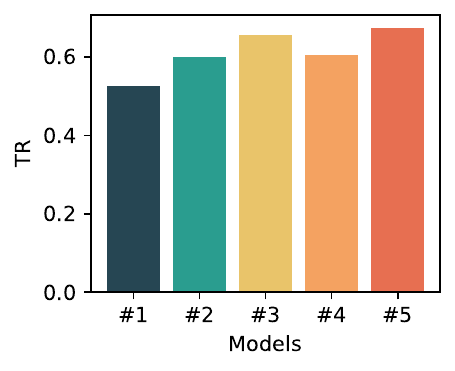}\label{fig: abl_tr}}
    \subfloat[MSE $\downarrow$]{\includegraphics[height=0.9in]{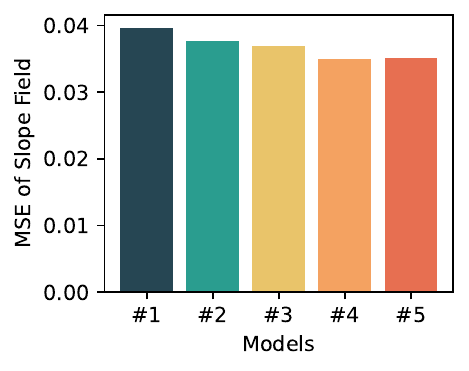}\label{fig: abl_mse}}
    \caption{Ablation Test for IFSN}
    \label{fig: abl_test}
\end{figure}

The specific settings of the ablation model are illustrated in Fig. \ref{fig: abl_tab}. Fig. \ref{fig: abl_sem}-\ref{fig: abl_mse} present the test results of four ablation models compared to our model, evaluated using three proposed metrics. The results indicate that all the ablation models maintain a semblance superior to manual picking, suggesting that the majority of curvature can be effectively identified by these models. In terms of TR, our model significantly outperforms the other ablation models. Furthermore, the test results for MSE confirm that the RMOs selected by our method exhibit a more consistent trend compared to manual picking than those of the other ablation models.

Based on the experimental results presented above, we confirm the effectiveness of the key improvement in IFSN for the curvature picking task. Additionally, these findings offer valuable insights for related research.

\subsection{Analysis of Inference Hyperparameters}
\begin{figure}[!ht]
    \centering
    \subfloat[]{\includegraphics[height=1.0in]{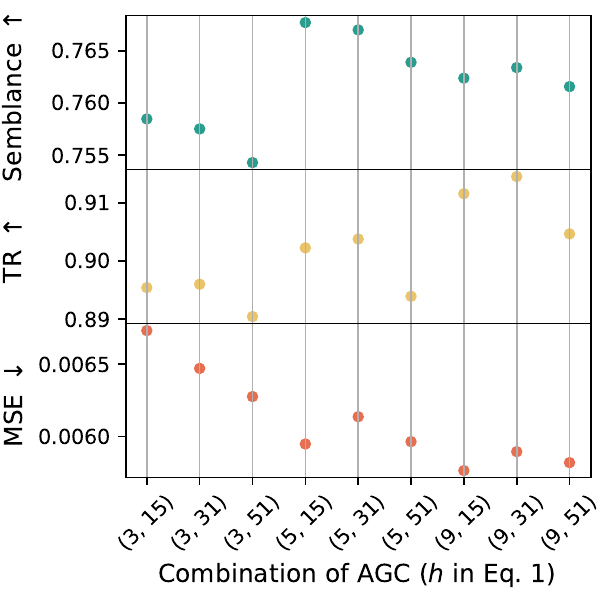}\label{fig: dis_agc}}
    \subfloat[]{\includegraphics[height=1.0in]{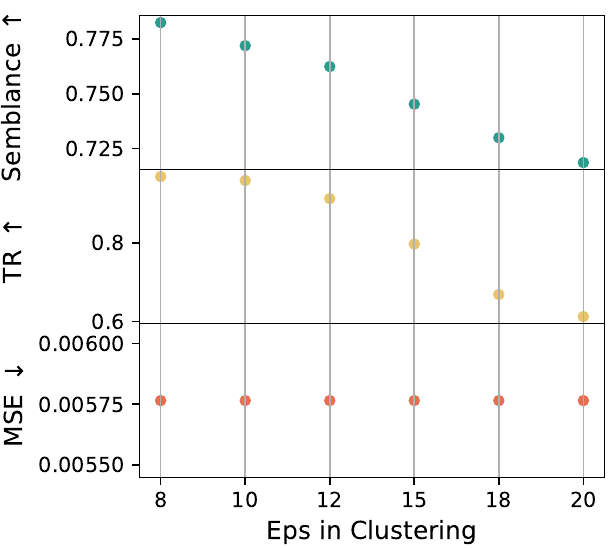}\label{fig: dis_clu_eps}}
    \subfloat[]{\includegraphics[height=1.0in]{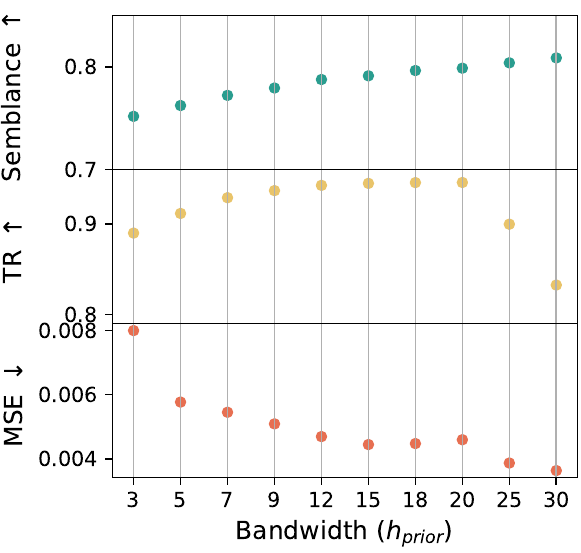}\label{fig: dis_win_k}}\\
    \subfloat[]{\includegraphics[height=1.0in]{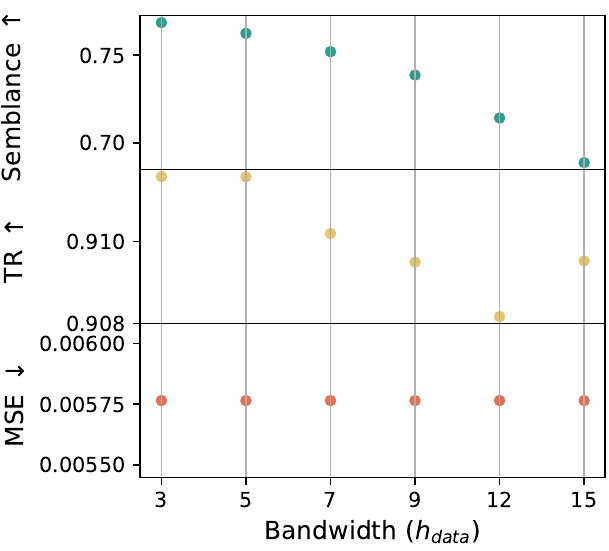}\label{fig: dis_h_data}}
    \subfloat[]{\includegraphics[height=1.0in]{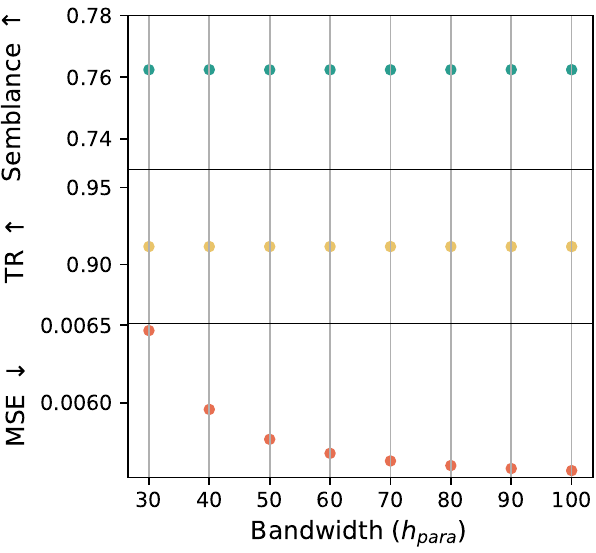}\label{fig: dis_h_para}}
    \subfloat[]{\includegraphics[height=1.0in]{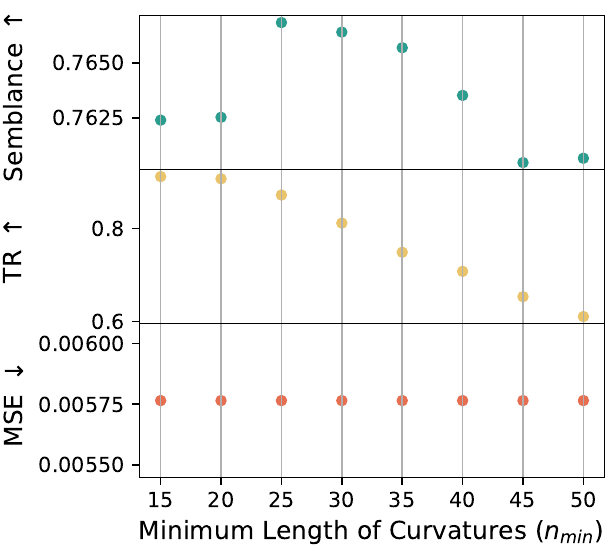}\label{fig: dis_min_len}}
    \caption{Sensitivity of Inference Hyperparameters}
    \label{fig: sensitivity}
\end{figure}

In the post-process of the cascade method, there are a few hyper-parameters controlling the curvature split, shown in Tab. \ref{tab: infer_para}. Aiming to learning a suitable selection method of these hyper-parameters, this section investigates the concrete effect of each hyper-parameter in terms of the BP dataset test. 
The sensitivity analysis results of each hyper-parameter is illustrated in Fig. \ref{fig: sensitivity}. 
Specifically, each test in Fig. \ref{fig: sensitivity} only adjust a inference hyper-parameter and maintain the others with the same setting as the combination in Tab. \ref{tab: infer_para}. 

Fig. \ref{fig: dis_agc} shows that for two hyper-parameter of the AGC features, the combination we used in Tab. \ref{tab: infer_para} are those that yield the lowest MSE, but may be not necessarily the optimal ones. When the AGC parameters are (9, 31), although the MSE increases slightly, the TR is higher. In scenarios where a higher TR is required, the latter choice might be more appropriate. Within the selected range of hyperparameters (Fig. \ref{fig: dis_agc}), the fluctuation in these three metrics is relatively large, indicating that this parameter needs to be manually adjusted in advance based on the characteristics of the dataset. 
The clustering parameter $d^{\text{eps}}$ determines the compactness of the clusters. A smaller $d^{\text{eps}}$ leads to tighter clustering, where only points that are very close to each other are grouped into the same cluster. Fig. \ref{fig: dis_clu_eps} also indicates that a larger $d^{\text{eps}}$ results in a decrease in TR. We analyze that this is because a larger $d^{\text{eps}}$ merges more curvatures, leading to sparse picking and thus a decline in TR. Therefore, we consider a smaller $d^{\text{eps}}$ to be a more conservative choice. For example, $d^{\text{eps}} = 8$ is a reasonable selection.
The parameters $h^{\text{prior}}$, $h^{\text{data}}$, and $h^{\text{para}}$ are key parameters controlling the Bayesian robust regression method. Concretely, the larger the parameter $h^{\text{prior}}$, the more neighborhoods will be calculated when computing the slope field. Although this enhances robustness, establishing connections with overly distant regions can also make the results excessively smooth. The results in Fig. \ref{fig: dis_win_k} also confirm this hypothesis: moderately increasing the parameter $h^{\text{prior}}$ can reduce the MSE and increase the TR, but when $h^{\text{prior}}$ is too large, the TR will also decrease.
The parameter $h^{\text{data}}$ controls the bandwidth during kernel regression. When its value is excessively large, the computational load will increase. Fig. \ref{fig: dis_h_data} indicates that the change in test accuracy is not significant, implying that the model is not sensitive to this parameter.
The parameter $h^{\text{para}}$ controls the range of the slope field considered. While a larger value can make the results more robust, it also leads to a greater computational burden. Fig. \ref{fig: dis_h_para} demonstrates that increasing the parameter $h^{\text{para}}$ can indeed reduce the MSE. However, considering the inference efficiency, we do not recommend choosing an excessively large value for this parameter.
The parameter $n_{\text{min}}$ controls the completeness requirement of the final picking. The larger its value, the more short curvatures will be removed. Fig. \ref{fig: dis_min_len} shows that as the parameter $n_{\text{min}}$ increases, the TR decreases correspondingly. When it is necessary to retain more complete curvatures, this parameter can be appropriately increased.

\section{Conclusion}
In this paper, we propose a high-precision RMO picking method for tomography inversion, which employs a cascade technique to identify RMOs and sequentially separate each RMO. After analyzing various experiments, we can draw four conclusions.
(1) The synthetic dataset generation method we proposed is effective, and the IFSN trained on this dataset can accurately identify the RMOs in the field data.
(2) The enhanced IFSN, based on the U-Net architecture, effectively captures curvature features in field data and demonstrates strong generalization capabilities.
(3) The proposed cascade method not only tracks curvatures with a higher degree of complexity but also ensures robustness and continuity.
(4) Compared to a semblance-based method and manual picking, the proposed cascade method achieves higher density and finer picking without requiring manual intervention.

\section*{Acknowledgment}
Thanks Hemang Shah and BP Exploration Operation Company Limited for providing the BP model dataset, which is downloaded in \url{https://wiki.seg.org/wiki/2007_BP_Anisotropic_Velocity_Benchmark}.

\ifCLASSOPTIONcaptionsoff
  \newpage
\fi

\bibliographystyle{IEEEtran}
\bibliography{reference}

\begin{thebibliography}{10}
\providecommand{\url}[1]{#1}
\csname url@samestyle\endcsname
\providecommand{\newblock}{\relax}
\providecommand{\bibinfo}[2]{#2}
\providecommand{\BIBentrySTDinterwordspacing}{\spaceskip=0pt\relax}
\providecommand{\BIBentryALTinterwordstretchfactor}{4}
\providecommand{\BIBentryALTinterwordspacing}{\spaceskip=\fontdimen2\font plus
\BIBentryALTinterwordstretchfactor\fontdimen3\font minus
  \fontdimen4\font\relax}
\providecommand{\BIBforeignlanguage}[2]{{%
\expandafter\ifx\csname l@#1\endcsname\relax
\typeout{** WARNING: IEEEtran.bst: No hyphenation pattern has been}%
\typeout{** loaded for the language `#1'. Using the pattern for}%
\typeout{** the default language instead.}%
\else
\language=\csname l@#1\endcsname
\fi
#2}}
\providecommand{\BIBdecl}{\relax}
\BIBdecl

\bibitem{brzostowski19923}
M.~A. Brzostowski and G.~A. McMechan, ``{3-D} tomographic imaging of
  near-surface seismic velocity and attenuation,'' \emph{Geophysics}, vol.~57,
  no.~3, pp. 396--403, 1992.

\bibitem{dou2012rockburst}
L.~Dou, T.~Chen, S.~Gong, H.~He, and S.~Zhang, ``Rockburst hazard determination
  by using computed tomography technology in deep workface,'' \emph{Safety
  Science}, vol.~50, no.~4, pp. 736--740, 2012.

\bibitem{Colin2026groundwater}
C.~A. Zelt, A.~Azaria, and A.~Levander, ``{3D} seismic refraction traveltime
  tomography at a groundwater contamination site,'' \emph{GEOPHYSICS}, vol.~71,
  no.~5, pp. H67--H78, 2006.

\bibitem{wang2017mechanism}
C.~Wang, A.~Cao, G.~Zhu, G.~Jing, J.~Li, and T.~Chen, ``Mechanism of rock burst
  induced by fault slip in an island coal panel and hazard assessment using
  seismic tomography: a case study from xuzhuang colliery, xuzhou, china,''
  \emph{Geosciences Journal}, vol.~21, pp. 469--481, 2017.

\bibitem{woodward1998automated}
M.~Woodward, P.~Farmer, D.~Nichols, and S.~Charles, ``Automated 3d tomographic
  velocity analysis of residual moveout in prestack depth migrated common image
  point gathers,'' in \emph{SEG Technical Program Expanded Abstracts
  1998}.\hskip 1em plus 0.5em minus 0.4em\relax Society of Exploration
  Geophysicists, 1998, pp. 1218--1221.

\bibitem{adler2008nonlinear}
F.~Adler, R.~Baina, M.~A. Soudani, P.~Cardon, and J.-B. Richard, ``Nonlinear 3d
  tomographic least-squares inversion of residual moveout in kirchhoff
  prestack-depth-migration common-image gathers,'' \emph{Geophysics}, vol.~73,
  no.~5, pp. VE13--VE23, 2008.

\bibitem{AlYahy1989semblance}
K.~Al-Yahya, ``Velocity analysis by iterative profile migration,''
  \emph{GEOPHYSICS}, vol.~54, no.~6, pp. 718--729, 1989.

\bibitem{Audebert1997scan}
F.~Audebert, J.~P. Diet, P.~Guillaume, I.~F. Jones, and X.~Zhang,
  ``{CRP‐scans}: {3D} {PreSDM} velocity analysis via zero‐offset
  tomographic inversion,'' in \emph{SEG Technical Program Expanded Abstracts
  1997}, 1997, pp. 1805--1808.

\bibitem{Woodward1998scan}
M.~Woodward, P.~Farmer, D.~Nichols, and S.~Charles, ``Automated {3D}
  tomographic velocity analysis of residual moveout in prestack depth migrated
  common image point gathers,'' in \emph{SEG Technical Program Expanded
  Abstracts 1998}, 1998, pp. 1218--1221.

\bibitem{Adler1999}
F.~Adler and S.~Brandwood, ``Robust estimation of dense 3d stacking velocities
  from automated picking,'' in \emph{SEG Technical Program Expanded Abstracts
  1999}, 1999, pp. 1162--1165.

\bibitem{siliqi2003high}
R.~Siliqi, D.~Le~Meur, F.~Gamar, L.~Smith, J.~Tour{\'e}, and P.~Herrmann,
  ``High-density moveout parameter fields v and $\eta$. part one: Simultaneous
  automatic picking,'' in \emph{SEG Technical Program Expanded Abstracts
  2003}.\hskip 1em plus 0.5em minus 0.4em\relax Society of Exploration
  Geophysicists, 2003, pp. 2088--2091.

\bibitem{siliqi2007high}
R.~Siliqi, P.~Herrmann, A.~Prescott, and L.~Capar, ``High order rmo picking
  using uncorrelated parameters,'' in \emph{SEG International Exposition and
  Annual Meeting}.\hskip 1em plus 0.5em minus 0.4em\relax SEG, 2007, p.
  2772–2776.

\bibitem{zhang2014automatic}
J.~Zhang, Q.~Yang, X.~Meng, and J.~Li, ``Automatic rmo picking in seismic
  travel time tomography,'' in \emph{2014 10th International Conference on
  Natural Computation (ICNC)}.\hskip 1em plus 0.5em minus 0.4em\relax IEEE,
  2014, pp. 1116--1120.

\bibitem{xu2024high}
J.~Xu and J.~Zhang, ``High-order residual moveout correction with global
  optimization in local time windows,'' \emph{Journal of Applied Geophysics},
  vol. 225, p. 105395, 2024.

\bibitem{Liu2010rtm}
J.~Liu and W.~Han, ``Automatic event picking and tomography on {3D RTM} angle
  gathers,'' in \emph{SEG Technical Program Expanded Abstracts 2010}, 2010, pp.
  4263--4268.

\bibitem{fruehn2014resolving}
J.~Fruehn, S.~Greenwood, V.~Valler, and D.~Sekulic, ``Resolving small-scale
  near-seabed velocity anomalies using non-parametric autopicking and hybrid
  tomography,'' \emph{CSEG Recorder}, vol.~39, no.~10, pp. 28--33, 2014.

\bibitem{Harris1999uncertainty}
P.~E. Harris and F.~Adler, ``Seismic resolution and uncertainty in time‐lapse
  studies,'' in \emph{SEG Technical Program Expanded Abstracts 1999}, 1999, pp.
  1671--1674.

\bibitem{Zhou2020mlpicking}
C.~Zhou and S.~Brown, ``Automatic velocity model building with machine
  learning,'' in \emph{SEG Technical Program Expanded Abstracts}, 2020, pp.
  1596--1600.

\bibitem{Bazargani2022segmentation}
F.~Bazargani, W.~Zhang, A.~Chandran, Z.~Liu, and H.~Rynja, ``Machine
  learning-based residual moveout picking,'' in \emph{SEG Technical Program
  Expanded Abstracts}, 2020, pp. 1674--1678.

\bibitem{Tschannen2020horizon}
V.~Tschannen, M.~Delescluse, N.~Ettrich, and J.~Keuper, ``Extracting horizon
  surfaces from {3D} seismic data using deep learning,'' \emph{GEOPHYSICS},
  vol.~85, no.~3, pp. N17--N26, 2020.

\bibitem{Xiong2018fault}
W.~Xiong, X.~Ji, Y.~Ma, Y.~Wang, N.~M. AlBinHassan, M.~N. Ali, and Y.~Luo,
  ``Seismic fault detection with convolutional neural network,''
  \emph{GEOPHYSICS}, vol.~83, no.~5, pp. O97--O103, 2018.

\bibitem{Wang2024FB}
H.~Wang, R.~Feng, L.~Wu, M.~Liu, Y.~Cui, C.~Zhang, and Z.~Guo, ``Dsu-net:
  Dynamic snake u-net for 2-d seismic first break picking,'' \emph{IEEE
  Transactions on Geoscience and Remote Sensing}, vol.~62, pp. 1--13, 2024.

\bibitem{Wu2019fault}
X.~Wu, L.~Liang, Y.~Shi, and S.~Fomel, ``{FaultSeg3D: Using synthetic data sets
  to train an end-to-end convolutional neural network for 3D seismic fault
  segmentation},'' \emph{GEOPHYSICS}, vol.~84, no.~3, pp. IM35--IM45, 2019.

\bibitem{Wu2023RL}
C.~Wu, B.~Feng, H.~Wang, X.~Song, S.~Sheng, and R.~Xu, ``An effective scheme
  for residual moveout picking using a markov decision process,'' \emph{IEEE
  Transactions on Geoscience and Remote Sensing}, vol.~61, p. Art. no. 5916311,
  2023.

\bibitem{Ma2018RL}
Y.~Ma and Y.~Luo, ``Automatic first-arrival picking with reinforcement
  learning,'' in \emph{SEG Global Meeting Abstracts}, 2018, pp. 493--497.

\bibitem{Ma2019RL}
Y.~Ma, T.~Fei, and Y.~Luo, ``A new insight into automatic first-arrival picking
  based on reinforcement learning,'' in \emph{81st EAGE Conference and
  Exhibition 2019}.\hskip 1em plus 0.5em minus 0.4em\relax European Association
  of Geoscientists \& Engineers, 2019, pp. 1--5.

\bibitem{woo2018cbam}
S.~Woo, J.~Park, J.-Y. Lee, and I.~S. Kweon, ``{CBAM}: {Convolutional} block
  attention module,'' in \emph{{Proceedings of the European Conference on
  Computer Vision (ECCV)}}, 2018, pp. 3--19.

\bibitem{ronneberger2015u}
O.~Ronneberger, P.~Fischer, and T.~Brox, ``{U-Net}: {Convolutional} networks
  for biomedical image segmentation,'' in \emph{{Medical Image Computing and
  Computer-Assisted Intervention--MICCAI 2015: 18th International Conference,
  Munich, Germany, October 5-9, 2015, Proceedings, Part III 18}}.\hskip 1em
  plus 0.5em minus 0.4em\relax Springer, 2015, pp. 234--241.

\bibitem{suzuki1985topological}
S.~Suzuki \emph{et~al.}, ``Topological structural analysis of digitized binary
  images by border following,'' \emph{Computer vision, graphics, and image
  processing}, vol.~30, no.~1, pp. 32--46, 1985.

\bibitem{ester1996density}
M.~Ester, H.-P. Kriegel, J.~Sander, X.~Xu \emph{et~al.}, ``A density-based
  algorithm for discovering clusters in large spatial databases with noise,''
  in \emph{{Proceedings of the Second International Conference on Knowledge
  Discovery and Data Mining (KDD'96)}}, vol.~96, no.~34, 1996, pp. 226--231.

\bibitem{ross2017focal}
T.-Y. Ross and G.~Doll{\'a}r, ``Focal loss for dense object detection,'' in
  \emph{Proceedings of the IEEE Conference on Computer Vision and Pattern
  Recognition}, 2017, pp. 2980--2988.

\bibitem{Loshchilov2017DecoupledWD}
I.~Loshchilov and F.~Hutter, ``Decoupled weight decay regularization,'' in
  \emph{International Conference on Learning Representations}, 2017.

\end{thebibliography}

\end{document}